\documentclass[5p]{elsarticle}

\usepackage{lineno,hyperref}
\usepackage{url}
\usepackage{amsmath}
\usepackage[lined,boxed,commentsnumbered,ruled]{algorithm2e} 
\usepackage{algorithmic}
\usepackage{amssymb}
\usepackage{array}
\newcolumntype{P}[1]{>{\centering\arraybackslash}p{#1}}
\newcolumntype{M}[1]{>{\centering\arraybackslash}m{#1}}
\usepackage{booktabs}
\usepackage{hyperref}
\usepackage{threeparttable}
\usepackage[table]{xcolor}
\usepackage{microtype}
\usepackage[labelfont=bf,singlelinecheck=false]{caption}
\usepackage{xcolor}
\usepackage{subcaption}

\journal{}
\biboptions{compress}

\DeclareMathOperator{\argmax}{arg\,max}

\begin{document}

\begin{frontmatter}

\title{Memetic Search for Vehicle Routing with Simultaneous Pickup-Delivery and Time Windows}

\author{Shengcai~Liu}
\ead{liusc3@sustech.edu.cn}
\author{Ke~Tang\corref{cor1}}
\ead{tangk3@sustech.edu.cn}
\author{Xin~Yao}
\ead{xiny@sustech.edu.cn}

\address{Guangdong Key Laboratory of Brain-Inspired Intelligent Computation,
Department of Computer Science and Engineering, Southern University of
Science and Technology, Shenzhen 518055, China}

\cortext[cor1]{Corresponding author}

\begin{abstract}
\textcolor{black}
{
The Vehicle Routing Problem with Simultaneous Pickup-Delivery and Time Windows (VRPSPDTW) has attracted much research interest in the last decade, due to its wide application in modern logistics.
Since VRPSPDTW is NP-hard and exact methods are only applicable to small-scale instances, heuristics and meta-heuristics are commonly adopted.
In this paper we propose a novel \textbf{M}emetic \textbf{A}lgorithm with efficien\textbf{T} local search and \textbf{E}xtended neighborhood, dubbed MATE, to solve this problem.
Compared to existing algorithms, the advantages of MATE lie in two aspects.
First, it is capable of more effectively exploring the search space, due to its novel initialization procedure, crossover and large-step-size operators.
Second, it is also more efficient in local exploitation, due to its sophisticated constant-time-complexity move evaluation mechanism.
Experimental results on public benchmarks show that MATE outperforms all the state-of-the-art algorithms, and notably, finds new best-known solutions on 12 instances (65 instances in total).
Moreover, a comprehensive ablation study is also conducted to show the effectiveness of the novel components integrated in MATE.
Finally, a new benchmark of large-scale instances, derived from a real-world application of the JD logistics, is introduced, which can serve as a new and more challenging test set for future research.
}

\end{abstract}


\end{frontmatter}


\section{Introduction}

Reverse logistics plays an important role in modern transportation.
Generally, it is related to bi-directional flow of goods regarding delivery and pickup activities, where the former refers to shipping goods to the customers, while the latter refers to the opposite.
Because of its significant effect on lowering costs associated with energy consumption and reducing
the environmental impact, reverse logistics has been incorporated into many regular delivery systems in various fields such as 
library books distribution \cite{MIN1989377},
grocery distribution \cite{Dethloff01},
and
parcel delivery \cite{BerbegliaCL10}.

In the literature, the problem involving bi-directional flow of goods has often been referred to as the pickup and delivery problem (PDP).
According to the surveys on PDP \cite{berbeglia2007static,BattarraCI14}, it can be further categorized into 3 different types:
1) many-to-many PDP where each commodity may have multiple origins and destinations and any location may be the origin or destination of multiple commodities;
2) one-to-many-to-one PDP where some commodities are delivered from a depot to many customers, while other commodities are collected at customers and delivered to the depot; 
3) one-to-one PDP where each commodity has a single origin and a single destination between which it must be delivered.
The most widely studied variant of the second type or one-to-many-to-one PDP, is the vehicle routing problem with simultaneous pickup and delivery (VRPSPD) \cite{MIN1989377,Koc2020}, due to the ever-growing trend toward recycling and product reuse.

This paper studies a practical situation of VRPSPD, which frequently occurs in modern logistics systems, e.g., JD logistics and CAINIAO logistics.
In these systems, in addition to delivering commodities purchased by customers from online, one must also plan the collection of used, defective, or obsolete products from customers;
moreover, in order to provide satisfactory service, either delivery or pickup needs to be operated within predefined time windows.
In the literature, this problem is referred to as the vehicle routing problem with simultaneous pickup-delivery and time windows (VRPSPDTW) \cite{WangC12}.

It has been shown that VRPSPDTW is NP-hard since it can be trivially reduced to VRPSPD (which is NP-hard) \cite{angelelli2002vehicle}.
Hence exact algorithms \cite{angelelli2002vehicle,WangC12} can only be used to find optimal solutions for small-scale instances (with number of customers smaller than 25).
As a result, researchers and practitioners are usually interested in developing meta-heuristics to find high-quality solutions within reasonable computational time.
The adopted search paradigms include
differential evolution \cite{mingyong2010improved},
genetic algorithm \cite{WangC12},
simulated annealing \cite{kassem2013solving,WangMZS15},
swarm intelligence optimization \cite{wulanhuang2016},
variable neighborhood search \cite{SHI2020103901},
and adaptive large neighborhood search \cite{HofS19}.

\textcolor{black}
{
However, despite the rich literature, in this area there still exist several important issues.
The first is that none of the current methods is able to perform both well and robustly across different types of problem instances.
In particular, on many practical instances, the solutions found by current methods are not satisfactory.
The root cause for this is that current methods are generally not very effective in either exploration or exploitation.
Typically they use a single insertion operator to construct initial solutions \cite{WangC12,kassem2013solving,WangMZS15} and conduct local search via traditional small step size move operators \cite{mingyong2010improved,WangC12,kassem2013solving,WangMZS15,wulanhuang2016,SHI2020103901}.
As a result, the diversity among the generated solutions is rather limited and the search process tends to get trapped in local optima, leading to unsatisfactory exploration.
Moreover, for existing methods the computational costs incurred by local search are generally very high \cite{WangC12,WangMZS15,wulanhuang2016,SHI2020103901}, which severely affects the efficiency of local exploitation.
The second issue is that little attention has been paid to large-scale instances.
Currently the only publicly available benchmark set in the literature \cite{WangC12} contains problem instances with customer number up to 100.
However, with the ever-growing of big cities, a real-world application might involve many more customers \cite{TangWLY17}.
Therefore, it is necessary to introduce a benchmark set containing large-scale real-world problem instances, to further facilitate the research in this area.
}

This paper is aimed at addressing the above issues.
Specifically, we propose a highly effective memetic algorithm (MA) to solve VRPSPDTW.
As an important area of evolutionary computation, MAs combine global search strategies (e.g., crossover) with local search heuristics, and have been shown to be very effective on a wide variety of combinatorial optimization problems \cite{DengW17,DuXZCH19,OsabaYFSLV19,TrachanatziRMM20}.
More importantly, MAs are the state-of-the-art approaches for many variants of the vehicle routing problems (VRP) \cite{TangMY09,MeiTY11,ZhangMTJ17,DecerleGHB19a,DecerleGHB19,chen2020heuristic,SabarBCTS19,SabarBCTS20,OkulewiczM19,WangWCCZX20,ChoongWL19,Goscien19}.
Since MA is a generic framework, appropriately instantiating it for a specific problem is non-trivial.
Generally, effective MAs should make good use of domain-specific knowledge in its main algorithm components (e.g., initialization, crossover operator and local search), and meanwhile achieve a good balance between exploration and exploitation.
The main contributions of this work can be summarized as follows.
\begin{enumerate}
  \item From the algorithmic perspective, the proposed MA involves several novel ingredients, which simultaneously enable effective exploration and efficient exploitation.
  First, we propose an initialization procedure by combining a construction heuristic with population-based search in an intelligent way, which can construct an initial population with high diversity.
  Second, we propose a new crossover operator for VRPSPDTW based on route inheritance and regret-based insertion.
  Third, we design a highly effective local search procedure for VRPSPDTW, which can flexibly search in a large neighborhood of a solution by switching between move operators with different step sizes.
  Moreover, we describe a sophisticated move evaluation process, which enables evaluating any neighborhood solution (issued from the move operators) in constant time, thus dramatically reducing the computational costs.
  Finally, we incorporate these components into the MA framework, and propose the \textbf{M}emetic \textbf{A}lgorithm with efficien\textbf{T} local search and \textbf{E}xtended neighborhood (MATE) for VRPSPDTW.
  \item From the computational perspective, MATE shows excellent performance on existing benchmark instances (65 instances in total). In particular, it outperforms all the four state-of-the-art algorithms. Notably, on 12 instances from the benchmark, MATE finds \textbf{new} best-known solutions.
  \item From the benchmarking perspective, we introduce a new instance set for VRPSPDTW, which can serve as a new and more challenging benchmark in this field.
  Compared to existing synthetic ones, the new instances are derived from a real-world application of JD logistics, and are with larger scales.
  The evaluation results of MATE on the new benchmark are also reported.
\end{enumerate}

The rest of the paper is organized as follows.
Section~\ref{sec:related_work} presents a literature review on the areas closely related to VRPSPDTW.
Section~\ref{sec:problem_description} formally defines the problem.
Section~\ref{sec:mate} first presents the framework of MATE, followed by its detailed implementation.
Section~\ref{sec:experiment} compares MATE with the state-of-the-art algorithms on the existing benchmark and introduces a new benchmark derived from a real-world application.
Finally, Section~\ref{sec:conclusion} concludes the paper.

\section{Related Work}
\label{sec:related_work}
As aforementioned, VRPSPDTW is a variant of VRPSPD.
The latter was first studied in \cite{MIN1989377} for a book distribution system involving 22 customers and two vehicles.
The solution was obtained by clustering customers into two groups and then solving the traveling salesman problem (TSP) for each group.
The results in \cite{MIN1989377} showed that, compared to traditional one-directional logistics, bi-directional logistics can achieve substantial time/distance savings.
Since then numerous studies have been conducted on VRPSPD.
The proposed approaches can be categorized into three groups:
exact approaches, heuristics and meta-heuristics.
There exist several exact algorithms for VRPSPD in the literature,
including branch-and-price algorithm \cite{DellAmicoRS06} with commodity-flow formulation,
branch-and-cut algorithm \cite{Rieck2013,SubramanianUPO11}
and branch-and-cut-and-price algorithm \cite{SubramanianUPO13}
with vehicle-flow formulation.

Compared to exact approaches, heuristics and meta-heuristics for solving VRPSPD have attracted much more research interest in the last decade, due to the fact that the problem is NP-hard \cite{angelelli2002vehicle}.
A number of construction heuristics,
tour partitioning \cite{montane2002vehicle},
parallel savings heuristic \cite{GajpalA10},
and residual capacity and radical surcharge (RCRS) heuristic \cite{Dethloff01},
have been proposed.
Compared to construction heuristics, meta-heuristics equipped with local search procedures can often obtain even better solutions.
Early research on meta-heuristics for solving VRPSPD mainly focused on tabu search (TS).
Various techniques, such as
record-to-record travel approximation \cite{ChenW06} and 
reactive mechanism \cite{WassanWN08}
have been proposed to enhance the performance of TS.
Later many other meta-heuristics have also been applied to solve VRPSPD, such as
iterated local search (ILS) \cite{subramanian2010parallel},
adaptive local search (ADL) \cite{AvciT15},
ant colony optimization (ACO) \cite{gajpal2009ant},
particle swarm optimization (PSO) \cite{AiK09},
and genetic algorithm (GA) \cite{VidalCGP14}.
One may refer to \cite{Koc2020} for a comprehensive review on existing approaches for VRPSPD.

VRPSPD with time-window constraints, i.e., VRPSPDTW, is probably the most studied variant of VRPSPD, due to its wide application in modern logistics.
The problem was first introduced in \cite{angelelli2002vehicle}, where the considered objective is to minimize the total travel distance (TD).
A branch-and-cut-and-price algorithm was also proposed.
Optimal solutions were obtained on small-scale instances with up to 20 customers.
Later the same objective was considered in \cite{mingyong2010improved,kassem2013solving}, in which a differential evolution (DE) approach and a simulated annealing (SA) approach were proposed, respectively.
The DE approach uses a decimal coding to construct an initial population and involves several new problem-specific move operators.
It was tested on instances with 8 and 40 customers, and the results were competitive.
The SA approach uses a sequential route construction heuristic to generate an initial solution, and then uses a SA procedure to improve the solution by searching in its neighborhood.
The evaluation results of the approach on instances with 10, 15 and 50 customers were also reported.

Another line of research in VRPSPDTW takes into account the number of used vehicles (NV), since the usage of a vehicle will result in its depreciation.
In particular, the primary goal is to minimize NV, and the second one is to minimize TD.
Such a setting was first considered in \cite{WangC12}, in which a co-evolutionary GA (Co-GA) was proposed.
Co-GA uses modified cheapest-insertion heuristics to generate initial solutions, and maintains two populations in the evolutionary process for diversification and intensification, respectively.
The well-known Solomon benchmark \cite{Solomon87} was modified in \cite{WangC12} to generate 65 instances with 10, 25, 50, and 100 customers.
Co-GA was then compared with the commercial solver CPLEX and the results showed the former can find better solutions within a comparatively shorter period of time.
Since then the benchmark introduced in \cite{WangC12} has become the most widely used benchmark in this area.
For convenience, we refer to this benchmark as \textit{wc} (authors' initials) set in this paper.

Following \cite{WangC12}, a number of approaches,
a parallel SA approach (p-SA) \cite{WangMZS15},
a swarm intelligence approach (IGAFSA) \cite{wulanhuang2016},
an adaptive large neighborhood search approach (ALNS-PR) \cite{HofS19}
and a two-stage hybrid approach (VNS-BSTS) \cite{SHI2020103901},
have been proposed to solve VRPSPDTW concerning minimizing NV and TD.
The SA approach adopts a slow cooling schedule and a randomized local search procedure.
It was tested on three 10-customer instances, three 25-customer instances, three 50-customer instances and six 100-customer instances from the \textit{wc} set.
On average, the solutions found by it were 0.22\% better (regarding TD) than Co-GA.
p-SA is a parallel variant of the SA approach.
It uses the RCRS heuristic to generate initial solutions and adopts a master–slave paradigm in which multiple SA procedures are run independently and simultaneously.
Competitive results on the \textit{wc} set were obtained, with 28 new best-known solutions in total (12 with lower NV and 16 with lower TD).
In addition, the evaluation results of p-SA on 30 large-scale instances, with customers of 200, 400, 600, 800 and 1000, were also reported.
However, the large-scale benchmark is not publicly available.
IGAFSA is a nature-inspired approach that allows infeasible solutions to enhance diversification.
On 39 instances from the \textit{wc} set, it found better solutions with lower TD than p-SA.
ALNS-PR uses seven removal operators and three reinsertion operators to repeatedly destroy and reconstruct a soluton, and involves a path-relinking component for intensification.
The results in \cite{HofS19} showed that ALNS-PR significantly outperformed all the previous approaches on the \textit{wc} set.
It found new best-known solutions for 48 instances in total (17 with lower NV and 31 with lower TD).
VNS-BSTS is the most recent approach for VRPSPDTW.
It has a two-stage solution framework.
In the first stage, it utilizes variable neighborhood search with a learning-based objective function to minimize NV, while in the second stage it uses tabu search to further minimize TD.
VNS-BSTS showed competitive performance on the \textit{wc} set, and found several new best-known solutions.




\section{Notations and Problem Definition}
\label{sec:problem_description}

Given a number of customers who require both pickup service and delivery service within certain time windows, the target of VRPSPDTW is to send out a fleet of capacitated vehicles, which are stationed at a depot, to meet the customer demands with the minimum total cost.
Formally, the problem is defined on a complete graph $G=(V,E)$ with $V=\{0,1,2,...,M\}$ as the node set and $E$ as the arc set defined between each pair of nodes, i.e., $E=\{\langle i,j \rangle|i,j \in V, i\neq j\}$.
For convenience, the depot is always denoted as 0 and the customers are denoted as $1,...,M$.
Each arc $\langle i,j \rangle \in E$ is associated with a travel distance $dist(i,j)$ and a travel time $time(i,j)$.
Each node $i \in V$ is associated with 5 attributes, i.e., a delivery demand $d_i$, a pickup demand $p_i$, a time window $[a_i, b_i]$ and a service time $s_i$.
$d_i$ is the amount of goods to deliver from the depot to customer $i$ and $p_i$ is the amount of goods to pick up from customer $i$ that must be delivered to the depot.
$a_i$ and $b_i$ are the start and the end of the time window in which the customer receives service.
Arrival of a vehicle at customer $i$ before $a_i$ results in a wait before service can begin; while arrival after $b_i$ is infeasible.
Finally, $s_i$ is the time spent by the vehicle to unload/load goods at customer $i$.
Note that for the depot, i.e., 0, $a_0$ and $b_0$ are the earliest time the vehicles can depart from the depot and the latest time the vehicles can return to the depot, respectively, and $d_0=p_0=s_0=0$. 

A fleet of $J$ identical vehicles, each with a capacity of $Q$ and a dispatching cost $u_1$, is initially located at the depot.
The vehicles depart from the depot and then serve the customers, and finally return to the depot.
Thus a solution $S$ to VRPSPDTW is represented by a set of vehicle routes, i.e., $S=\{R_1,R_2,...,R_K\}$, in which each route $R_i$ consists of a sequence of nodes that the vehicle visits, i.e., $R_i=(h_{i,1},h_{i,2},...,h_{i,L_i})$, where $h_{i,j}$ is the $j$-th node visited in $R_i$, and $L_i$ is the length of $R_i$.
For the sake of brevity, in the following we temporarily omit the subscript $i$ in $R_i$, i.e., $R=(h_1, h_2,...,h_L)$.
The total travel distance of $R$, denoted as $TD(R)$, is:
\begin{equation}
\label{eq:td}
TD(R) = \sum_{j=1}^{L-1} dist(h_j,h_{j+1}).
\end{equation}
The time of arrival at and the time of departure from $h_j$, denoted as $arr(h_j)$ and $dep(h_j)$, respectively, can be computed recursively via the following equations:
\begin{equation}
\begin{split}
&dep(h_1) = a_0\\
&arr(h_j) = dep(h_{j-1}) + time(h_{j-1}, h_j),\ j>1\\
&dep(h_j) = \mathrm{max} \left\{ arr(h_j), a_{h_j} \right\} + s_{h_j},\ j>1.
\end{split}
\end{equation}
The vehicle load on arrival at $h_j$, denoted as $load(h_j)$, is:
\begin{equation}
\begin{split}
&load(h_1) = \sum_{j=1}^{L}d_{h_j}\\
&load(h_j) =  load(h_{j-1}) - d_{h_{j-1}} + p_{h_{j-1}}, \ j>1.
\end{split}
\end{equation}



The total cost of $S$, denoted as $TC(S)$, consists of two parts: the dispatching cost of the used vehicles, which is $u_1 \cdot K$, and the transportation cost, which is the total travel distance of $S$ multiplied by cost per unit of travel distance $u_2$.
The objective is to find a $S$ with the minimum TC, as presented in Eq.~(\ref{eq:vrpspdtw}):
\begin{equation}
\label{eq:vrpspdtw}
\begin{split}
\min_{S}\ &TC(S) \triangleq u_1 \cdot K + u_2 \cdot \sum_{i=1}^K TD(R_i)\\
s.t.:\ & K \leq J\\
& h_{i,1}=h_{i,L_i}=0,\ 1 \leq i \leq K\\
& \sum_{i=1}^K \sum_{j=2}^{L_i-1} \mathbb I[h_{i,j} = x]=1,\ 1 \leq x \leq M\\
& load(h_{i,j}) \leq Q,\ 1 \leq i \leq K, 1 \leq j \leq L_i \\
& arr(h_{i,j}) \leq b_{h_{i,j}},\ 1 \leq i \leq K, 2 \leq j \leq L_i \\
& dep(h_{i,1}) \geq a_0\ \mathrm{and}\ arr(h_{i, L_i}) \leq b_0,\ 1 \leq i \leq K \\
\end{split},
\end{equation}
where the constraints are:
1) the number of used vehicles cannot exceed the number of available ones; 
2) each route must start from and return to the depot;
3) each customer must be served exactly once (note $\mathbb I[\cdot]$ is the indicator function); 4) vehicles cannot be overloaded during transportation;
5) service to each customer must be performed within that customer’s time window;
6) vehicles can only depart after the start of the time window of the depot ($a_0$) and must return to the depot before the end of its time window ($b_0$).
It is noted that the two different objectives considered in the literature of VRPSPDTW (see Section~\ref{sec:related_work}) are both special cases of Eq.~(\ref{eq:vrpspdtw}).
Specifically, the dispatching cost $u_1$ could be set to 0 to only minimize TD.
The ratio of $u_1$ to $u_2$, i.e., $u_1/u_2$, could be set to a sufficiently large number such that minimizing NV is the primary goal and minimizing TD is the second one.

\begin{figure}[t]
	\centering
	\scalebox{1.0}{\includegraphics[width=\linewidth]{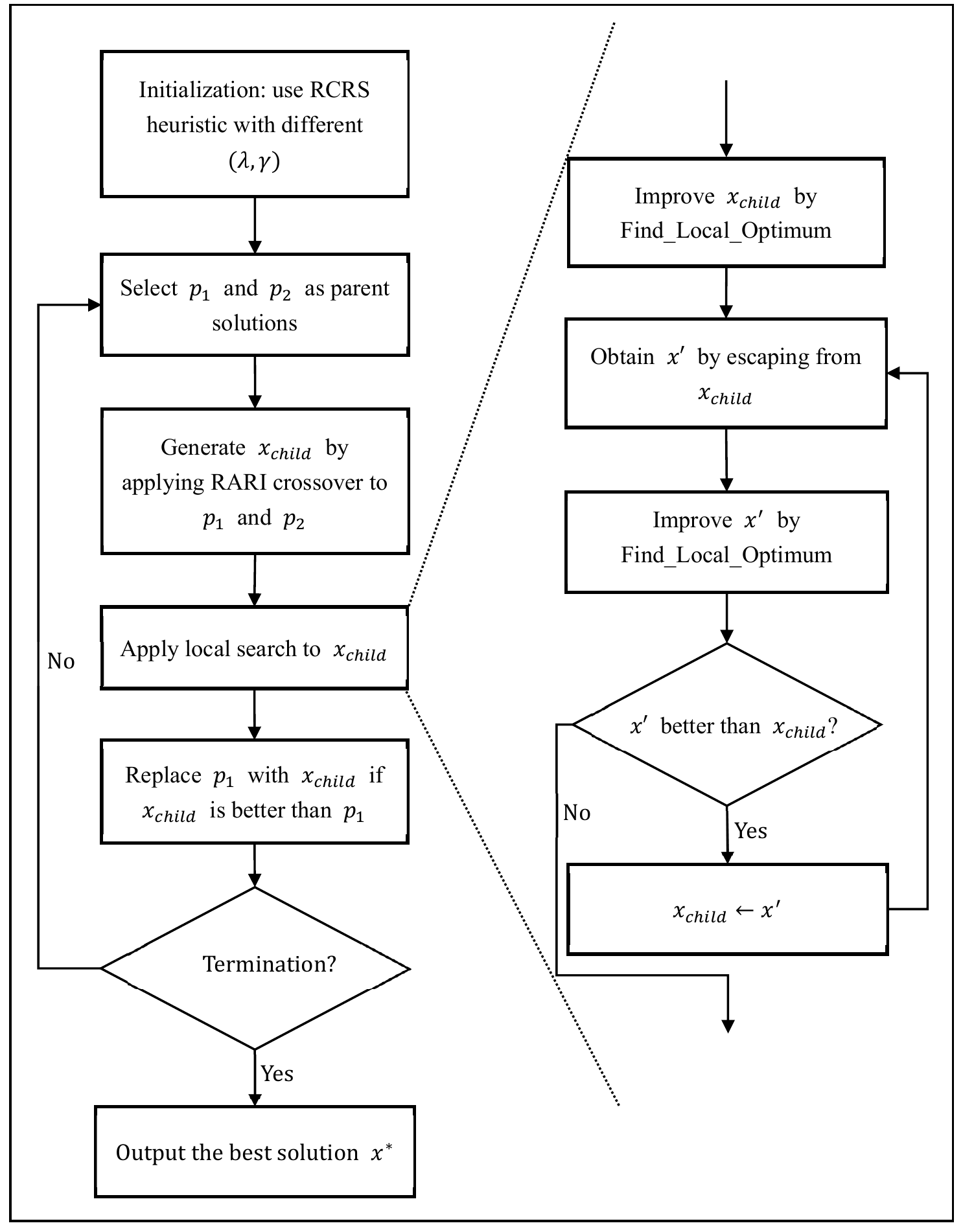}}
	\caption{\textcolor{black}{The flow chart of the proposed memetic algorithm.}}
	\label{fig:flow_chart}
\end{figure}

\textcolor{black}
{
The main characteristics of VRPSPDTW lie in the capacity aspect and the temporal aspect.
The former is that customers can simultaneously have delivery demand and pick-up demand.
Since these two types of demands have different effects on the vehicle load (one will increase the load while the other will decrease the load), such a characteristic causes difficulty assigning appropriate customers to vehicles under the capacity constraint.
The latter is that each customer is associated with a hard time window, which further increases the difficulty planning the service order for a particular vehicle under the time-window constraint.
In addition to taking the above constraints into account, considering the exponential search space of VRPSPDTW, it is also crucial for the algorithm to maintain sufficient exploration as well as efficient exploitation, to find high-quality feasible solutions.
}

\section{Memetic Algorithm for VRPSPDTW}
\label{sec:mate}
\textcolor{black}{
	In this section we first describe the general framework of MATE, and then elaborate on its three main components (initialization, crossover and local search) in turn.
	The flow chart of the algorithm is also illustrated in Figure~\ref{fig:flow_chart}.
	Note that these components ensure that the generated solutions always satisfy the constraints of capacity and time windows.
	Moreover, if a generated solution’s route number (i.e., the number of used vehicles) exceeds the number of available ones, then the solution will be immediately discarded.
	In other words, 
	during the entire run of MATE, each individual in the population always corresponds to a feasible solution to the VRPSPDTW instance, and its fitness is the inverse of its total cost as defined in Eq.~(\ref{eq:vrpspdtw}).
	}

\begin{algorithm}[tbp]
  \LinesNumbered
  \SetKwInOut{Input}{input}
  \SetKwInOut{Output}{output}
  \Input{A VRPSPDTW instance; population size, $N$; longest consecutive generations without improvement, $G_{max}$}
  \Output{the best found solution $x^*$}
  $\{x_1,x_2,...,x_{N}\} \gets$ $\mathrm{Initialization()}$;\\
  $x^* \gets \mathrm{Select\_Best}(x_1,x_2,...,x_{N})$;\\
  \Repeat{$x^*$ not improved in the last $G_{max}$ generations}
  {
    $\pi\left( \cdot \right) \gets$ a random permutation of $1,...,N$;\\
    \For{$i \gets 1 $ \KwTo $N$}
    {
      $p_1 \gets x_{\pi(i)},p_2 \gets x_{\pi(i+1)}$;\\
      $x_{child} \gets \mathrm{Crossover}(p_1, p_2)$;\\
      $x_{child} \gets \mathrm{Local\_Search}(x_{child})$;\\
      $x_{\pi(i)} \gets \mathrm{Select\_Best}(x_{child}, p_{1})$;\\
    }
    $x^* \gets \mathrm{Select\_Best}(x^*,x_1,x_2,...,x_{N})$;\\
  }
\Return{$x^*$}
\caption{The General Framework of MATE}
\label{alg:mate}
\end{algorithm}

\subsection{General Framework}
As presented in Algorithm~\ref{alg:mate}, the proposed algorithm, called MATE, combines population-based evolutionary search and local optimization.
More specifically, the population consists of $N$ individuals, where $N$ is a parameter.
After initialization (line 1), MATE enters an evolutionary process.
In each generation (lines 4-11), each individual in the population is selected once as parent $p_1$ and once as parent $p_2$, in a random order (lines 4 and 6).
Such a mechanism could promote the population diversity since each individual has exactly the same chance of being selected.
For each pair of parents, MATE generates an offspring solution via the crossover operator (line 7), and then tries to improve it by local search (line 8).
After that, the better one among the offspring solution and $p_1$ will replace the population member selected as $p_1$ (line 9).
Therefore, no replacement will take place if the offspring solution is worse than $p_1$.
In other words, the loss of population diversity is allowed only if the average fitness of the population is improved.
The iterations of generation in MATE terminate when the best found solution has not been improved (line 11) in the last consecutive $G_{max}$ generations (line 12), in which case the algorithm is considered to have converged.
Finally, the best found solution is returned (line 13).

\begin{algorithm}[tbp]
  \LinesNumbered
  \SetKwInOut{Input}{input}
  \SetKwInOut{Output}{output}
  \Input{population size, $N$}
  \Output{$\{x_1,x_2,...,x_N\}$}
  $count \gets 1$;\\
  \For{$i \gets 1$ \KwTo $\sqrt{N}$}
  {
    \For{$j \gets 1$ \KwTo $\sqrt{N}$}
    {
      $\lambda \gets \frac{1}{\sqrt{N}-1} \cdot (i-1)$;\\
      $\gamma \gets \frac{1}{\sqrt{N}-1} \cdot (j-1)$;\\
      $x_{count} \gets \mathrm{RCRS}(\lambda, \gamma)$;\\
      $count \gets count+1$;\\ 
    }
  }
\Return{$\{x_1,x_2,...,x_N\}$}
\caption{The Initialization Procedure}
\label{alg:initialization}
\end{algorithm}

\subsection{Initialization}
\label{sec:init}
The initial population is constructed using the RCRS heuristic, which is an extension of the cheapest-insertion heuristic.
A detailed description of the RCRS heuristic can be found in \cite{Dethloff01}.
\textcolor{black}
{Specifically, RCRS heuristic builds a route by iteratively inserting an unassigned customer into the route at the position with the minimal value w.r.t the RCRS criterion, until no feasible insertions (regarding constraints of capacity and time windows) exist.
If no feasible insertions exist, a new route is activated.
The above procedure is repeated until all customers are assigned.}
The core of the RCRS heuristic is the RCRS criterion, which assesses how good a potential insertion is.
The conventional cheapest-insertion heuristic uses the TD criterion, i.e., the extra travel distance caused by inserting a customer, which is usually considered short-sighted.
The RCRS criterion extends the TD criterion in two aspects.
First, the remaining vehicle capacity after an insertion is taken into account, i.e., the residual capacity (RC) criterion, which measures the degrees of freedom for future insertions.
Second, the distances of customers to the depot are considered, i.e., the radial surcharge (RS) criterion, which seeks to avoid the unfavorable extra travel distances caused by inserting the remotely located customers which are ``left over'' to a late stage of the insertion procedure.
RCRS criterion is a combination of RC criterion and RS criterion, with two weighting parameters $\lambda,\gamma \in [0,1]$.
Unfortunately, the best values of $\lambda$ and $\gamma$ vary across different problem instances.
For a given instance it is hard to determine a good choice of $\lambda$ and $\gamma$ in advance.

However, in MATE we can take advantage of the population based search by trying different values of $\lambda$ and $\gamma$ for different individuals.
As presented in Algorithm~\ref{alg:initialization}, for each of $\lambda$ and $\gamma$, we start with the value of 0, and gradually increase it with a step size of $1/(\sqrt{N}-1)$ until it reaches 1 (lines 4-5).
In summary, for each of $\lambda$ and $\gamma$, we use $\sqrt{N}$ different values ranging from 0 to 1 that are equally spaced, thus obtaining a total of $N$ different combinations of $(\lambda, \gamma)$, based on each of which the RCRS heuristic is used to construct an initial solution (line 6).
Finally, it is required that the parameter $N$, i.e., the population size, is a square number., e.g., 4, 9, 16, etc.

\textcolor{black}
{
Compared to previous methods \cite{Dethloff01,WangMZS15} that also use the RCRS heuristic to construct initial solutions, the proposed initialization approach has at least two-fold advantages.
First, it eliminates the need of tedious tuning of the two parameters $\lambda$ and $\gamma$ of the RCRS heuristic, since it simultaneously tries very different values of $\lambda$ and $\gamma$ for different initial individuals.
Second, it enables good coverage on the design space of ($\lambda, \gamma$), which is beneficial to constructing an initial population with high diversity, thus promoting more sufficient exploration in the search space.
The effect of the proposed initialization procedure on MATE's performance is also investigated in the experiments (see Section~\ref{sec:ablation_study}).
}

\begin{algorithm}[tbp]
  \LinesNumbered
  \SetKwInOut{Input}{input}
  \SetKwInOut{Output}{output}
  \Input{parent solutions $p_1,p_2$}
  \Output{$x_{child}$}
  $x_{child} \gets$ initialize an empty solution with no routes;\\
  \Repeat
  {no more inherited routes are feasible}
  {
    copy random route from $p_1$ to $x_{child}$;\\
    copy random route from $p_2$ to $x_{child}$;\\
  }
  \tcc{--- regret-based insertion ---}
  $U \gets $ all the remaining unassigned customers;\\
  \While{$U \neq \varnothing$}
  {
    \ForEach{node $v \in U$}
    {
      calculate $regret(v)$ according to Eq.~(\ref{eq:regret});\\
    }
    $v^* \gets \argmax_{v \in U} regret(v)$;\\
    Insert $v^*$ into $x_{child}$ at its best insertion position;\\
    $U \gets U \backslash \{v^*\}$;\\
  }
\Return{$x_{child}$}
\caption{The RARI Crossover Operator}
\label{alg:crossover}
\end{algorithm}

\subsection{Crossover Operator}
\label{sec:crossover}
An effective crossover operator should be able to transmit useful building blocks from parents to offspring, and meanwhile incorporate domain-specific heuristics to generate high-quality offspring that are significantly different from either of the parents.
In this paper, we propose a new route-assembly-regret-insertion (RARI) crossover.

As presented in Algorithm~\ref{alg:crossover}, RARI crossover first repeatedly chooses a random route from each parent solution in turns to insert into the offspring solution until no feasible insertion exists (lines 2-5).
In other words, the offspring solution will inherit as many routes as possible from each parent equally.  
After that, the remaining unassigned customers (if there exist) will be inserted into the offspring solution to form a complete solution.
In the literature of VRP, one commonly used procedure to insert these nodes is the two-step insertion \cite{AlvarengaMT07}, which first tries inserting them into existing routes of the offspring solution and then inserts the remaining nodes by cheapest-insertion heuristic.
\textcolor{black}
{Here, we propose to use regret-based insertion (lines 7-15) to insert the unassigned customers.
Such an approach uses a regret value, which represents the expected cost of inserting a node not in this iteration but in a future iteration, to assess how good a potential insertion is.
Compared to cheapest-insertion which is usually considered short-sighted, regret-based insertion can avoid the postponing issue of cheapest-insertion heuristic---placing ``difficult'' nodes (which are expensive to insert) late in the process where there exist few opportunities for inserting them as many of the routes are already ``full''.
Moreover, compared to two-stage insertion, regret-based insertion is a simpler one-stage approach that simultaneously considers both the insertion of customers into existing routes and new routes.
The comparison between two-stage insertion and regret-based insertion is also conducted in the experiments (see Section~\ref{sec:ablation_study}).}

Specifically, the regret value for a node $v$ is obtained as follows.
Supposing that there exist $m$ routes in the current partial solution, i.e., $\{R_1,R_2,...,R_m\}$,
for each route $R_i$, the position and the cost $c_i$ of the \textcolor{black}{best feasible insertion} (based on the objective value) of $v$ into $R_i$ is calculated.
Moreover, the cost of inserting $v$ into an empty route, i.e., a route from the depot to $v$ and back to the depot, is also calculated, which is denoted as $c_0$.
Then $c_0,c_1,...,c_m$ are sorted in ascending order, and the sorting results are denoted as $c_{\pi(0)},c_{\pi(1)},...,c_{\pi(m)}$, i.e., $c_{\pi(0)} \leq c_{\pi(1)}\leq ... \leq c_{\pi(m)}$.
The regret value of $v$, denoted as $regeret(v)$, is the difference in the cost of inserting $v$ in its best route and its second best route:
\begin{equation}
	\label{eq:regret}
	regret(v) = c_{\pi(1)} - c_{\pi(0)}.
\end{equation}
\textcolor{black}{In each iteration of the regret-based insertion (lines 9-14), first the regret value of each currently unassigned customer is calculated (lines 9-11), and then the one with the maximum regret value (ties are broken by selecting the insertion with the lowest cost) will be inserted at its best feasible position (lines 12-14)}.
Informally speaking, in each iteration the insertion that we will regret most if it is not done now will be carried out.
The iterations terminate when all unassigned customers have been inserted into the offspring solution.

\captionsetup[sub]{font=small}
\begin{figure*}[t]
	\centering
	\begin{subfigure}[b]{1.0\columnwidth}
		\centering
		\scalebox{1.0}{\includegraphics[width=\linewidth]{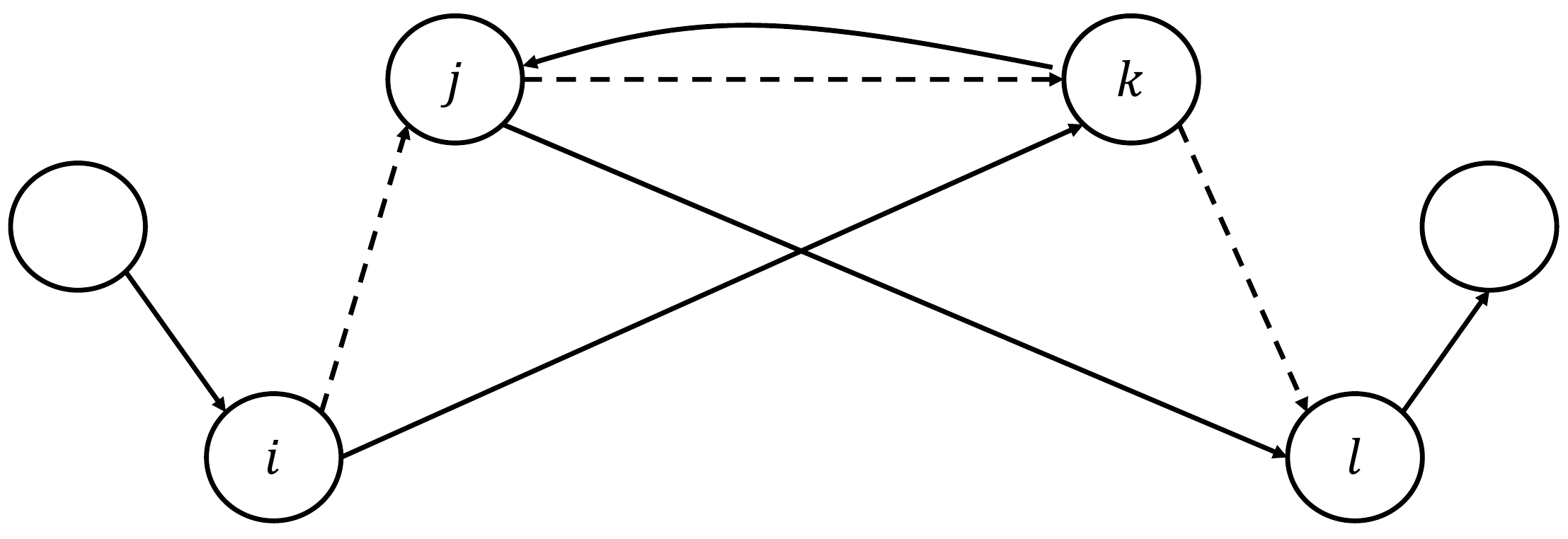}}
		\caption{\textcolor{black}{\textit{2-opt.} Inverting a subsequence of two consecutive customers $j$ and $k$ in a route.}}
		\label{fig:move_2-opt}
	\end{subfigure}
	\hfill
	\begin{subfigure}[b]{1.0\columnwidth}
		\centering
		\scalebox{1.0}{\includegraphics[width=\linewidth]{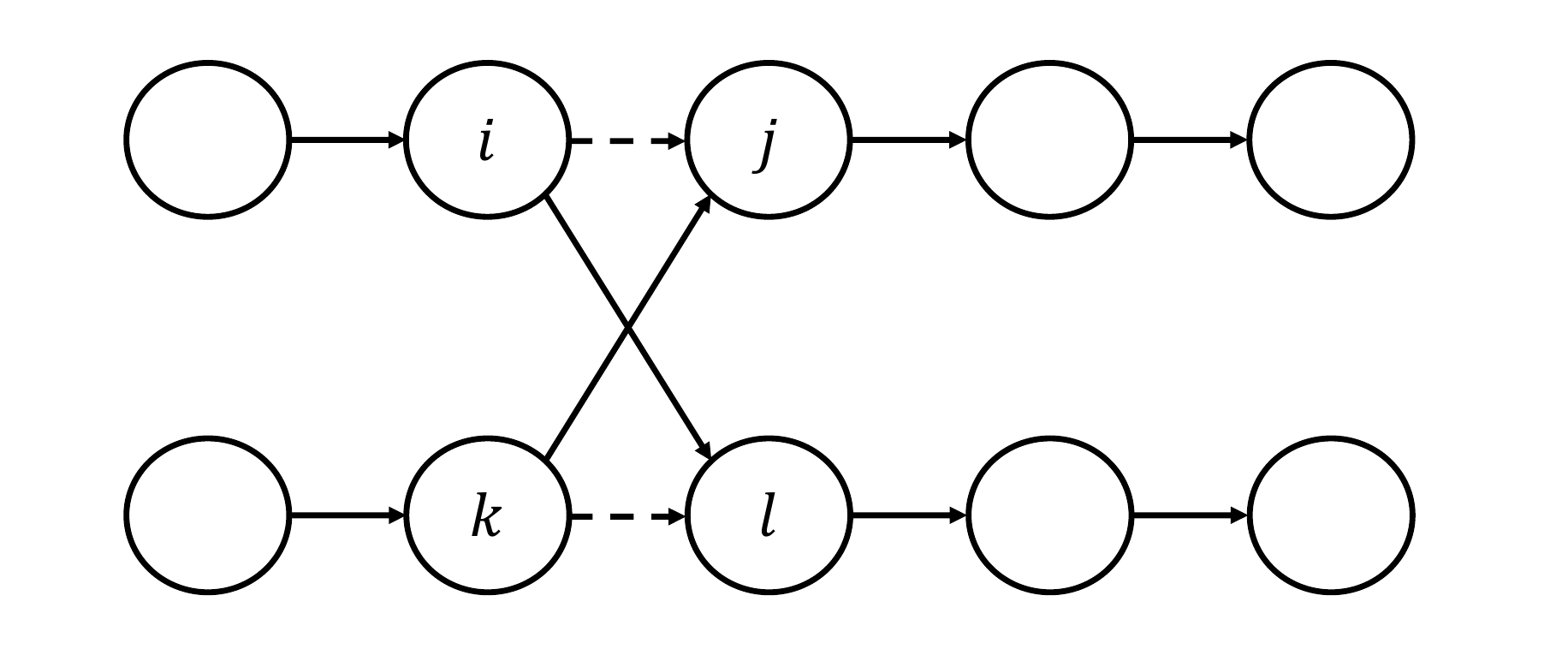}}
		\caption{\textcolor{black}{\textit{2-opt*.} Reconnecting the first part of one route at node $i$ with the second part of another route at node $l$ and vice versa.}}
	\end{subfigure}
	\begin{subfigure}[b]{1.0\columnwidth}
		\centering
		\scalebox{1.0}{\includegraphics[width=\linewidth]{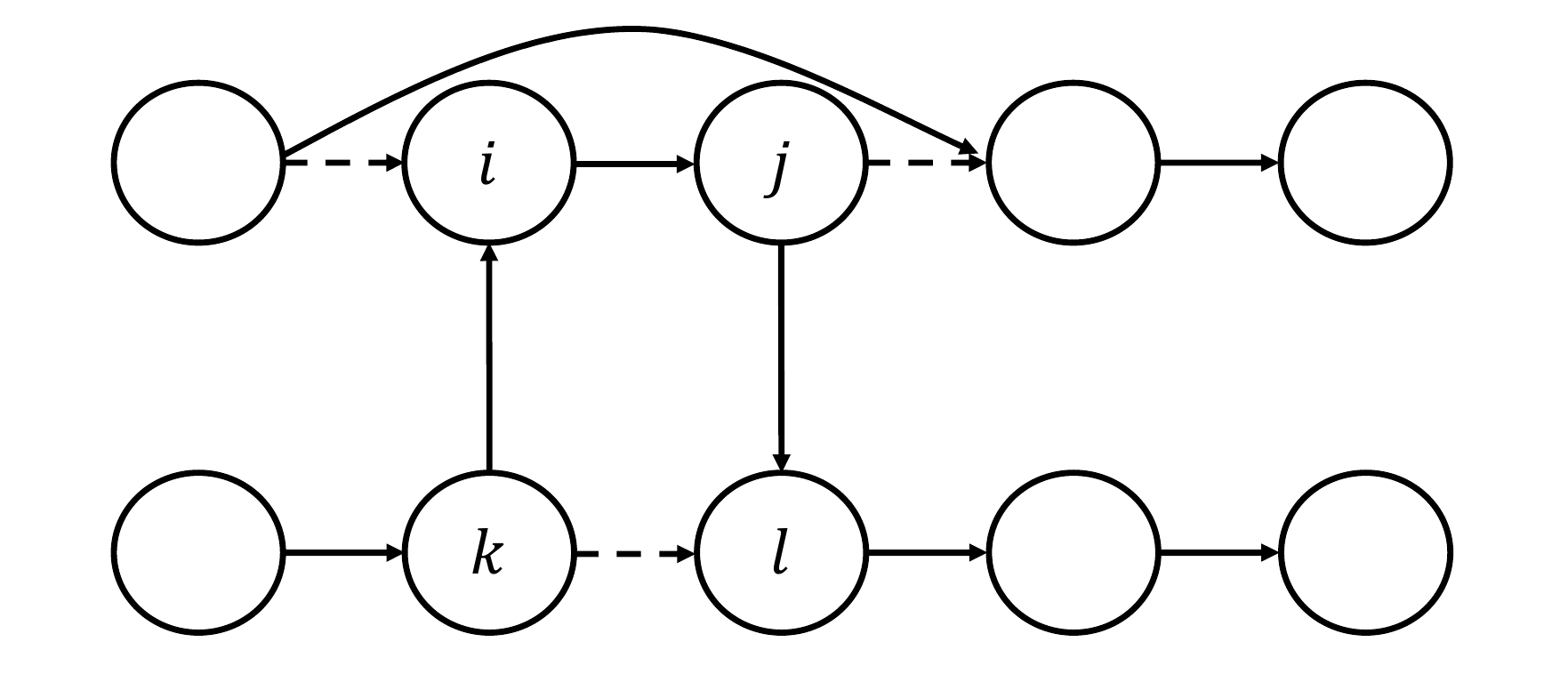}}
		\caption{\textcolor{black}{\textit{or-opt.} Removing a subsequence of two consecutive customers $i$ and $j$ from a route, and then reinserting it between $k$ and $l$ in another route.}}
	\end{subfigure}
	\hfill
	\begin{subfigure}[b]{1.0\columnwidth}
		\centering
		\scalebox{1.0}{\includegraphics[width=\linewidth]{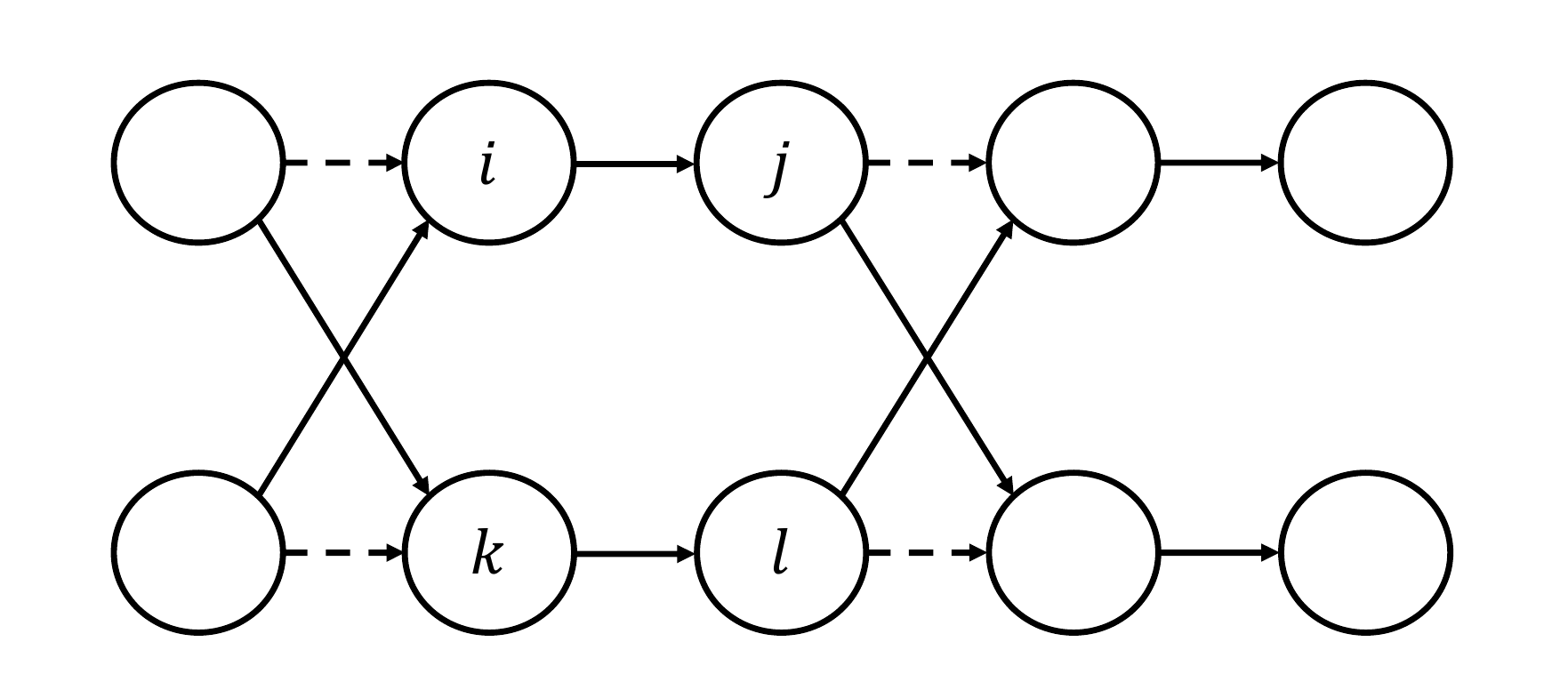}}
		\caption{\textcolor{black}{\textit{swap.} Exchanging a subsequence of two customers $i$ and $j$ in a route with a subsequence of two customers $k$ and $l$ in another route.}}
	\end{subfigure}
	\caption{\textcolor{black}{Illustrative examples of all move operators used. The dashed lines and the solid lines indicate the removed arcs and the new arcs in the routes due to performing the move, respectively.}}
	\label{fig:moves}
\end{figure*}

\subsection{Local Search Procedure}
\label{sec:local_search}
The local search procedure is the core component of MATE.
Its effectiveness has a decisive impact on the performance of the algorithm.
\textcolor{black}
{
Existing VRPSPDTW algorithms \cite{WangC12,WangMZS15,wulanhuang2016} typically conduct local search via some traditional move operators, which modify only a small part of the current solution.
In other words, these operators have small search step sizes which are expected to perform well in small search space.
However, for difficult problems with large search spaces and many local optima, they may perform unsatisfactorily.
Moreover, for existing algorithms the computational costs incurred by local search are very high, often accounting for the major part of the total costs.
As a result, there is large room for further reducing the computational complexity of local search.
To address these issues, we propose a novel local search procedure which can simultaneously achieve the following three goals.
First, it can quickly identify high-quality local optima in a small search space.
Second, it can jump out from the current local optimum to other promising regions.
Third, its computational complexity is very low such that the incurred computational cost is reasonable.
}

To accomplish the first two goals, the local search procedure switches between move operators with different step sizes.
As demonstrated in Algorithm~\ref{alg:local_search}, given a solution $x$, it first finds the local optimum around $x$ in a small region defined by several traditional move operators (line 1).
Then, it extends the search step size by applying the removal-and-reinsertion operator (lines 3-7), trying to jump out from the local optimum.
After that, the step size will be reduced again to identify the local optimum $x'$ in the new local region (line 9).
If $x'$ is better than the current solution $x$, the latter will be updated and then used as the base solution for the next iteration (line 10).
Otherwise the procedure will terminate (line 11) and the current solution $x$ is returned (line 13).

\subsubsection{Finding Local Optimum}
To identify the local optimum around a given solution, the embedded $\mathrm{Find\_Local\_Optimum}$ sub-procedure in Algorithm~\ref{alg:local_search} utilizes several traditional move operators \cite{BraysyG05}, \textcolor{black}{namely \textit{2-opt}, \textit{2-opt*}, \textit{or-opt} and \textit{swap}, as illustrated in Figure~\ref{fig:moves}.}
The \textit{2-opt} operator inverts a subsequence of two consecutive customers in a route.
The \textit{2-opt*} operator removes two arcs from two different routes to divide each route into two parts, and then reconnects the first part of the first route with the second part of the second route and vice versa.
The \textit{or-opt} operator removes a subsequence of one or two consecutive customers from a route, and then reinserts it into another position of the same route or a different route.
The \textit{swap} operator exchanges two subsequences of one or two consecutive customers, which may be on the same route (but not overlapping with each other), or on different routes. 

Given a solution $x$, a best-improvement search is conducted in the neighborhoods of $x$ defined by the four above-described move operators.
That is, all solutions that can be reached by applying either of the four operators to $x$ are evaluated, and \textcolor{black}{the best feasible solution} among them, say $\bar{x}$, is then compared with $x$.
If $\bar{x}$ is better than $x$, the latter will be updated.
This procedure is repeated until no further improvement can be found.
By this means, it is ensured that a local optimum has been reached in each neighborhood w.r.t the move operators.

\begin{algorithm}[tbp]
	\LinesNumbered
	\SetKwInOut{Input}{input}
	\SetKwInOut{Output}{output}
	\Input{solution $x$; the lower and upper bounds for the proportion of the removed nodes, $\omega_1, \omega_2$}
	\Output{$x$}
	$x \gets \mathrm{Find\_Local\_Optimum}(x)$;\\
	\While{true}
	{
		$x' \gets x$;\\
		\tcc{escape from local optimum}
		$q \gets $ a random number $\in [\omega_1 M, \omega_2 M]$;\\
		$U \gets $ remove $q$ nodes from $x'$;\\
		Insert $U$ into $x'$ via regret-based insertion;\\
		\tcc{exploit the new local region}
		$x' \gets \mathrm{Find\_Local\_Optimum}(x')$;\\
		\lIf{$x'$ is better than $x$}{$x \gets x'$}
		\lElse{break}
	}
	\Return{$x$}
	\caption{The Local Search Procedure}
	\label{alg:local_search}
\end{algorithm}

\subsubsection{Escaping from Local Optimum}
The above-described move operators modify only a small part (one or two routes) of the solution;
thus the neighborhoods defined by them are actually a small region around the solution.
Once a local optimum has been found in the region, an operator with a bigger step size is needed for jumping out of it \cite{Yao91,yao1992dynamic,TangMY09}.
We propose to use a removal-and-reinsertion operator to accomplish this goal. 
This operator first removes a number of customers from the solution (lines 5-6 in Algorithm~\ref{alg:local_search}), and then reinserts them into the solution again (line 7 in Algorithm~\ref{alg:local_search}).
The number of the removed nodes, i.e., $q$, is a random number in $[\omega_1 M, \omega_2 M]$, where $M$ is the total number of customers and $\omega_1,\omega_2$ are two parameters satisfying $0< \omega_1 < \omega_2 <1$.
In this paper, $\omega_1,\omega_2$ are set to 0.2 and 0.4, respectively.
In other words, 20-40\% of all customers in the solution are rearranged by the removal-and-reinsertion operator; thus it is very likely that the obtained new solution will be quite different from the original one.

The node removal procedure seeks to remove customers that are correlated.
More specifically, it first randomly removes a customer.
Then two customers that are most correlated to the last removed customer are identified.
After that, roulette wheel selection is used to select one of these two customers to remove.
This procedure is repeated until $q$ customers have been removed in total.
The correlation of a customer $j$ to a customer $i$ is the weighted sum of the distance, the minimum waiting time, and the minimum time-window violation on a direct service from $i$ to 
$j$:

\small
\begin{equation}
\scriptscriptstyle
\begin{split}
\label{eq_correlation}
corr(i,j) = dist(i,j) + \eta \cdot [ & \max\{a_j - time(i,j) - s_i - b_i, 0\}\\
		   + \gamma \cdot & \max\{ a_i + s_i + time(i,j) - b_j, 0\} ],
\end{split}
\end{equation}
\normalsize
where $\eta$ is a factor that rescales time into distance and equals to the average distance between all nodes divided by the average travel time between all nodes.
$\gamma$ is a positive penalty factor (i.e., 10) for the time-window violation.
The lower $corr(i, j)$ is, customer $j$ is more correlated to customer $i$.
Intrinsically, Eq.~(\ref{eq_correlation}) measures how good customer $j$ is as the next serviced one after customer $i$.
Therefore it is expected to be reasonably easy to shuffle those highly-correlated customers (which have been removed from the current solution) around and thereby create perhaps new better solutions.
After node removal, the node reinsertion is conducted once again using the regret-based insertion heuristic (see lines 6-15 in Algorithm~\ref{alg:crossover}), \textcolor{black}{ensuring the final solution is feasible}.

\subsubsection{Move Evaluation in Constant Time}
To accomplish the third goal of the local search procedure, i.e., keeping its computational complexity as low as possible, we propose to optimize the time complexity of move evaluation, i.e., calculating the cost and checking the feasibility of the solutions issued from the move operators (\textit{2-opt}, \textit{2-opt*}, \textit{or-opt} and \textit{swap}), since it is the most time-consuming part of the local search procedure with a huge number of solutions being generated and evaluated.
Recall that these move operators will modify at most two routes of a solution.
Hence, to evaluate a move, one only needs to evaluate the changed routes, which could be trivially done in $O(n)$ by traversing the routes, where $n$ is the route length.
In this paper, we propose a new move evaluation approach for VRPSPDTW which has constant time complexity, i.e., $O(1)$.

\begin{figure}[t]
	\centering
	\scalebox{1.0}{\includegraphics[width=\linewidth]{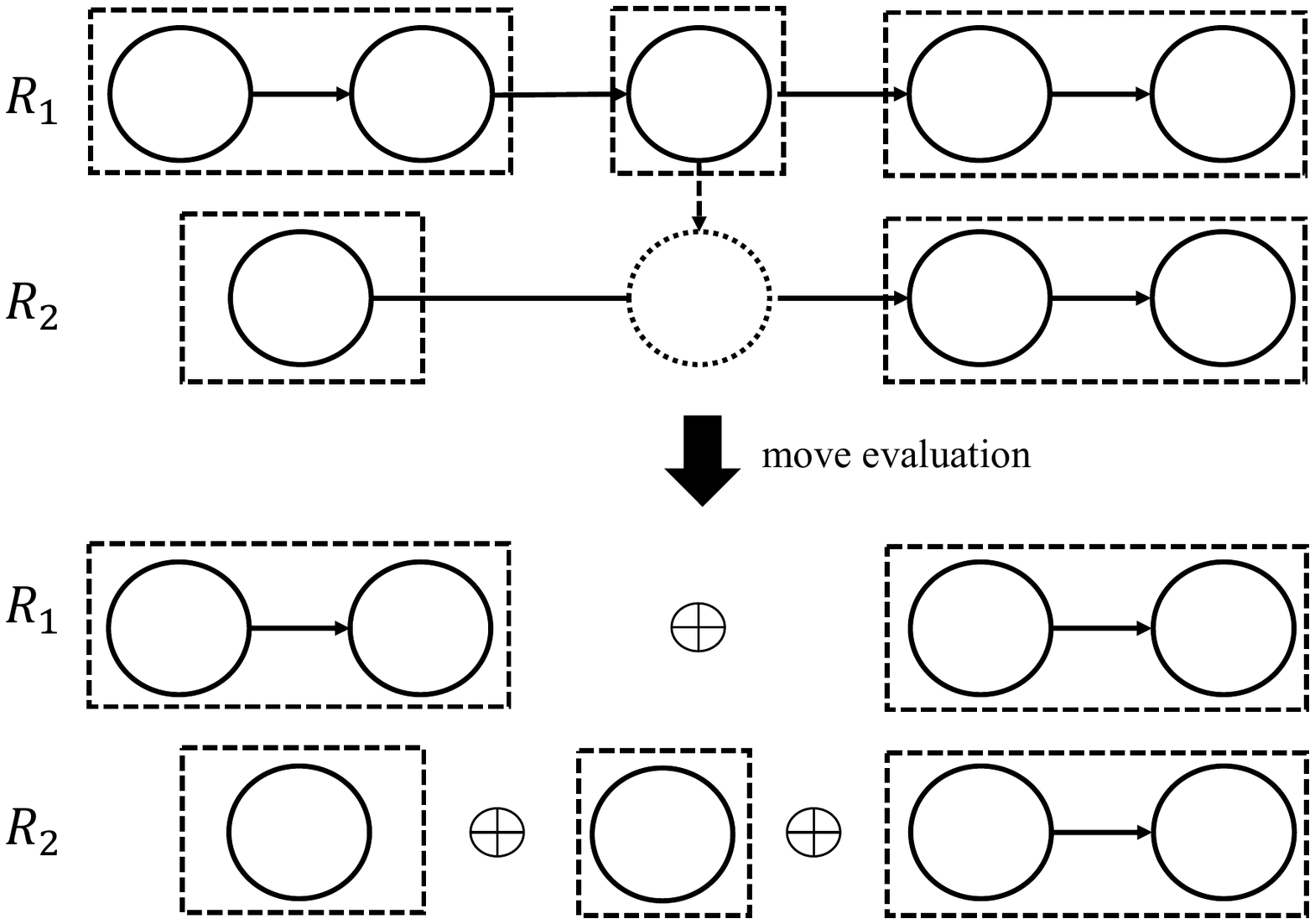}}
	\caption{\textcolor{black}{An illustrative example of evaluating a \textit{or-opt} move, which relocates a node from route $R_1$ to route $R_2$, using data of precomputed subsequences (dashed boxes) and the concatenation operator $\oplus$.}}
	\label{fig:move_eval}
\end{figure}
\textcolor{black}
{
The basic idea of sequence-based evaluation is that any move can be viewed as a separation of routes into subsequences, which are then concatenated into new routes.
As illustrated in Figure~\ref{fig:move_eval}, supposing the \textit{or-opt} operator relocates a node from route $R_1$ to route $R_2$ (the upper part of the figure), the two new routes $R_1$ and $R_2$ issued from this move can be evaluated using concatenation of the subsequences (the lower part of the figure).
Let $\sigma$ and $\oplus$ denote a subsequence and the concatenation operator, respectively.
Let $\sigma^0$ denote a subsequence involving a single node $i$.
For each route in a solution, we can recursively compute the following attributes of its all subsequences by concatenation.}
\paragraph{Distance}
\textcolor{black}{
The travel distance of $\sigma^0$, $W(\sigma^0)$, is always 0.
Moreover, the travel distance of a concatenation of two subsequences $\sigma_1$ and $\sigma_2$ is the sum of $W(\sigma_1)$ and $W(\sigma_2)$ plus the distance from the last node of $\sigma_1$ (denoted by $j$) to the first node of $\sigma_2$ (denoted by $k$).}
\begin{equation}
\begin{split}
	&W(\sigma^0) = 0\\
	&W(\sigma_1 \oplus \sigma_2) = W(\sigma_1) + W(\sigma_2) + dist(j,k).
\end{split}
\end{equation}
\paragraph{Capacity}
\textcolor{black}{
Let $C_{E}(\sigma)$, $C_{H}(\sigma)$ and $C_{L}(\sigma)$ denote the initial, the highest, and the final loads on the vehicle during $\sigma$, respectively.
For $\sigma^0$, its initial load and final load are the delivery demand $d_i$ and the pick-up demand $p_i$, respectively;
its highest load is the maximum of these two.
For a concatenation of $\sigma_1$ and $\sigma_2$, its initial load is the sum of the initial loads of $\sigma_1$ and $\sigma_2$, i.e., the total delivery demands of $\sigma_1$ and $\sigma_2$; its final load is the sum of the final loads of $\sigma_1$ and $\sigma_2$, i.e., the total pick-up demands of $\sigma_1$ and $\sigma_2$.
The highest load during $\sigma_1 \oplus \sigma_2$ is the maximum of the highest loads during $\sigma_1$ and $\sigma_2$, when concatenating $\sigma_1$ and $\sigma_2$.}
\begin{equation}
\begin{split}
	&C_E(\sigma^0)=d_i\\
	&C_L(\sigma^0)=p_i\\
	&C_H(\sigma^0)=\max\{C_E(\sigma^0), C_L(\sigma^0)\}\\
	&C_E(\sigma_1 \oplus \sigma_2) = C_E(\sigma_1) + C_E(\sigma_2)\\
	&C_L(\sigma_1 \oplus \sigma_2) = C_L(\sigma_1) + C_L(\sigma_2)\\
	&C_H(\sigma_1 \oplus \sigma_2) = \max\{C_H(\sigma_1) +C_E(\sigma_2),\\
	&\ \ \ \ \ \ \ \ \ \ \ \ \ \ \ \ \ \ \ \ \ \ \ \ \ \ \ \ \ \ \ \ C_L(\sigma_1)+C_H(\sigma_2) \}.
\end{split}
\end{equation}

\paragraph{Time Window}
\textcolor{black}{
Let $D(\sigma)$ denote the minimum duration on $\sigma$.
For the arrival time to the first node of $\sigma$, there exists an interval $[E(\sigma), L(\sigma)]$ which allows the minimum duration.
For $\sigma^0$, its minimum duration is the service time $s_i$, while $E(\sigma)$ and $L(\sigma)$ are the start and the end of the time windows of the node of $\sigma^0$.
For a concatenation of $\sigma_1$ and $\sigma_2$, the following equations enable computing the same data on the concatenation of two subsequences $\sigma_1$ and $\sigma_2$ (let $j$ and $k$ denote the first node of $\sigma_1$ and the last node of $\sigma_2$, respectively).}
\begin{equation}
\begin{split}
&D(\sigma^0)=s_i\\
&E(\sigma^0)=a_i\\
&L(\sigma^0)=b_i\\
&\Delta=D(\sigma_1) + time(j,k)\\
&\Delta_{WT}=\max\{E(\sigma_2) - \Delta -L(\sigma_1), 0\}\\
&\Delta_{TW}=E(\sigma_1)+D(\sigma_1)+time(i,j)-L(\sigma_2)\\
&D(\sigma_1 \oplus \sigma_2)=D(\sigma_1) + D(\sigma_2) + time(j,k) + \Delta_{WT}\\
&E(\sigma_1 \oplus \sigma_2)=\max \{E(\sigma_2) -\Delta, E(\sigma_1) \} - \Delta_{WT}\\
&L(\sigma_1 \oplus \sigma_2) = \min \{ L(\sigma_2) - \Delta, L(\sigma_1)\}.
\end{split}
\end{equation}
\paragraph{Move Evaluation}
\textcolor{black}
{
The sequence-based evaluation approach is used in the best-improvement search in the neighborhoods of solution $S$, i.e., the $\mathrm{Find\_Local\_Optimum}$ sub-procedure.
Specifically, the attributes of all the subsequences (and their reversal) of $S$ are precomputed using the above equations, and will be used for checking the feasibility and computing the costs of routes issued from the move operators.
Supposing a route $R$ is generated by applying a move operator and $R=\sigma_1 \oplus \sigma_2 \oplus \cdots \sigma_t$.
We first retrieve the precomputed attributes of $\sigma_1,...,\sigma_t$, and then calculate the corresponding attributes of $R$ by applying the above equations.
The feasibility of $R$ can be checked in the following way:
1) $R$ is feasible regarding time-window constraint if $\Delta_{TW} \leq 0$;
2) $R$ is feasible regarding capacity constraint if $C_H \leq C$, where $C$ is the vehicle capacity.
Moreover, we use the total travel distance $W$ to calculate the transportation cost of $R$, $u_2 \cdot W$, where $u_2$ is the cost per unit of travel distance.
The four move operators used in this paper correspond to a concatenation of less than five subsequences, i.e., $t \leq 5$.
Hence, given the data on subsequences, any move evaluation is performed in constant time.
Note that if $S$ is updated, the attributes of the changed routes need to be recomputed in $O(n^2)$, where $n$ is the route length.
However, this is acceptable since in the search process the number of updates is smaller in orders of magnitude than the number of move evaluation.
}

\subsection{Discussion}
\textcolor{black}
{It is worth mentioning that the novel components of MATE, i.e., the initialization, crossover and local search procedures, are also applicable to other VRP variants.
The reason is that the core components of these procedures, i.e., the RCRS heuristic, the route-inheritance heuristic equipped with regret-based insertion, and the move operators with different step sizes, can be smoothly used as long as the problem being solved involves capacitated vehicles and concerns minimizing the total travel distance, which is very common in many VRP variants.
Moreover, the efficient move evaluation approach can also be used to accelerate existing VRPSPDTW algorithms, since it is applicable to all classical move operators (not just the ones used in this paper) which correspond to the concatenation of finite number of subsequences.}

\section{Computational Study}
\label{sec:experiment}
To evaluate the effectiveness of MATE, we compared it with a number of state-of-the-art algorithms on existing benchmark set and a new benchmark set derived from a real-world application.
After that, we conducted a comprehensive ablation study to assess the contribution of the novel components integrated in MATE.
Finally, the effect of different values of the parameter $N$ (the population size) was investigated.
The source code of MATE, as well as all the benchmark sets used in the experiments, are made at \url{https://github.com/senshineL/VRPenstein}.

\subsection{Benchmark Sets}
\label{sec:benchmark}
The commonly used (also the only publicly available) benchmark set in the literature of VRPSPDTW, referred to as the \textit{wc} set in this paper (see Section~\ref{sec:related_work}), was generated by \cite{WangC12} through modifying the well-known Solomon benchmark \cite{Solomon87}.
The \textit{wc} set contains 65 instances in total, including 9 small-scale instances (three 10-customer instances, three 25-customer instances, and three 50-customer instances) and 56 medium-scale instances (100-customer instances).
Note all these instances are defined in the two-dimensional Euclidean space; that is, each node $i$ (customer or depot) has a coordinate $(x_i,y_i)$ and the distance between two nodes $i$ and $j$ is the Euclidean distance, i.e., $\sqrt{(x_i-x_j)^2 + (y_i-y_j)^2}$.
This is a special case of the problem formulation given in this paper (see Section~\ref{sec:problem_description}), where the distances between nodes are explicitly given no matter whether they are defined in the Euclidean plane.
According to the distribution of the customers' locations as well as the intensity of the time-window constraints and the capacity constraints, instances in the \textit{wc} set could be further categorized into 6 types (subsets), namely Cdp1{\raise.17ex\hbox{$\scriptstyle\sim$}}, Cdp2{\raise.17ex\hbox{$\scriptstyle\sim$}}, Rdp1{\raise.17ex\hbox{$\scriptstyle\sim$}}, Rdp2{\raise.17ex\hbox{$\scriptstyle\sim$}}, RCdp1{\raise.17ex\hbox{$\scriptstyle\sim$}} and RCdp2{\raise.17ex\hbox{$\scriptstyle\sim$}}, where the categories refer to:
\begin{enumerate}
\item Cdp: customers' locations are clustered;
\item Rdp: customers' locations are uniform randomly distributed;
\item RCdp: customers' locations are a mix of random and clustered locations;
\item Type 1: customers have narrow time windows and vehicle's capacity is small;
\item Type 2: customers have large time windows and vehicle's capacity is large.
\end{enumerate}
For all instances in the \textit{wc} set, minimizing NV (number of used vehicles) is the primary goal and minimizing TD (total distance) is the second one.
To meet this, for the objective function Eq.~(\ref{eq:vrpspdtw}), the ratio of the dispatching cost of each vehicle ($u_1$) to the cost per unit of travel distance ($u_2$), i.e., $u_1/u_2$, should be set to a sufficiently large number (see Section~\ref{sec:problem_description}).
Following \cite{WangC12}, in the experiments we set $u_1$ and $u_2$ to 2000 and 1, respectively.

As aforementioned, the \textit{wc} set contains only synthetic instances of small or medium scales.
To further assess the potential of the algorithms, we introduce a new benchmark set derived from the distribution system of JD logistics.
In this system, in addition to delivering the goods purchased by customers, one also needs to collect goods (e.g., defective goods or goods in need of maintenance) from customers, and both of these operations must be executed in predefined time windows to provide satisfactory service. 
Therefore in essence it could be modeled as the VRPSPDTW problem considered here.
Further, original data was collected from this system, which contains requests occurring during a period of time in a city.
The total number of requests is 3000.
We then sampled from the data to generate instances with 200, 400, 600, 800, and 1000 customers.
For each problem scale, we generated 4 instances, which finally gave us a benchmark set of 20 instances.
The new benchmark set is called the \textit{jd} set.
Unlike the \textit{wc} set, for this new set there is no priority for minimizing either NV or TD.
Instead, $u_1$ and $u_2$ are directly given based on the estimated values in the application, and the objective is to minimize TC (total cost), i.e., the sum of the dispatching cost and the transportation cost, as defined in Eq.~(\ref{eq:vrpspdtw}).


\subsection{Compared Algorithms and Algorithm Settings}

\begin{table}[tbp]
	\centering
	\caption{Algorithm Settings of MATE. ``$M$'' is the number of the customers.}
	\label{tab:alg_setting}
	\scalebox{1.0}{
		\begin{tabular}{M{.1\columnwidth}M{.43\columnwidth}M{.3\columnwidth}}
			\toprule
			& Description & Value \\
			\midrule
			$G_{max}$ & longest consecutive generations without improvement & 50 \\
			\midrule
			$\omega_1, \omega_2$ & lower and upper bounds for the
			proportion of the removed nodes & $0.2, 0.4$ \\
			\midrule
			$N$ & population size & 
			\[
			\begin{cases}
				16, & M \leq 100\\
				36, & M > 100
			\end{cases}
			\] \\
			\bottomrule
	\end{tabular}}
\end{table}

\begin{table*}[t]
	\centering
	\caption{Comparison between CPLEX, Co-GA, and the proposed MATE on small-scale instances in the \textit{wc} set.
		For each instance, the best performance regarding TC (i.e., $2000 \cdot \text{NV}+\text{TD}$) is highlited in grey. ``$M$'' is the number of customers of the instance.
		``Time'' refers to the rescaled computation time in seconds.
		``$^a$'' indicates the ``out-of-memory'' values.
		``Avg$\pm$std'' is the average and the standard deviation of NV and TD of the 30 solutions found by the 30 independent runs.
		``Best'' is the best solution regarding TC among the 30 solutions found by the 30 independent runs.}
	\scalebox{0.994}{
		\begin{tabular}{llllllllllll}
			\toprule
			\multicolumn{1}{l}{Instance$\vert$\textit{M}} & \multicolumn{3}{l}{CPLEX} & \multicolumn{3}{l}{Co-GA}  & \multicolumn{5}{l}{MATE} \\
			\multicolumn{1}{r}{} &       &       &       &       &       &       &    Avg$\pm$std &       & Best  &  &     \\
			\cmidrule(lr){2-4}
			\cmidrule(lr){5-7}
			\cmidrule(lr){8-9}
			\cmidrule(lr){10-11}
			\cmidrule(lr){12-12}
			\multicolumn{1}{r}{} & NV    & TD    & Time     & NV    & TD    & Time    & NV    & TD    & NV    & TD    & Time   \\
			\midrule
			RCdp1001$\vert$10 & \cellcolor{gray!50}3     & \cellcolor{gray!50}348.98 & 1      & \cellcolor{gray!50}3     & \cellcolor{gray!50}348.98 & 1   & \cellcolor{gray!50}3.00$\pm$0.00   & \cellcolor{gray!50}348.98$\pm$0.00 & \cellcolor{gray!50}3     & \cellcolor{gray!50}348.98 & 1 \\
			RCdp1004$\vert$10 & \cellcolor{gray!50}2     & \cellcolor{gray!50}216.69 & 1387   & \cellcolor{gray!50}2     & \cellcolor{gray!50}216.69 & 1   & \cellcolor{gray!50}2.00$\pm$0.00   & \cellcolor{gray!50}216.69$\pm$0.00 & \cellcolor{gray!50}2     & \cellcolor{gray!50}216.69 & 1 \\
			RCdp1007$\vert$10 & \cellcolor{gray!50}2     & \cellcolor{gray!50}310.81 & 23     & \cellcolor{gray!50}2     & \cellcolor{gray!50}310.81 & 1    & \cellcolor{gray!50}2.00$\pm$0.00   & \cellcolor{gray!50}310.81$\pm$0.00 & \cellcolor{gray!50}2     & \cellcolor{gray!50}310.81 & 1 \\
			RCdp2501$\vert$10 & \cellcolor{gray!50}5     & \cellcolor{gray!50}551.05 & 15     & \cellcolor{gray!50}5     & \cellcolor{gray!50}551.05 & 3   & \cellcolor{gray!50}5.00$\pm$0.00   & \cellcolor{gray!50}551.05$\pm$0.00 & \cellcolor{gray!50}5     & \cellcolor{gray!50}551.05 & 1 \\
			RCdp2504$\vert$25 & $7^a$     & $738.32^a$ & 448193  & \cellcolor{gray!50}4     & \cellcolor{gray!50}473.46 & 2   & \cellcolor{gray!50}4.00$\pm$0.00   & \cellcolor{gray!50}473.46$\pm$0.00 & \cellcolor{gray!50}4     & \cellcolor{gray!50}473.46 & 1 \\
			RCdp2507$\vert$25 & $7^a$     & $634.20^a$ & 405429  & \cellcolor{gray!50}5     & \cellcolor{gray!50}540.87 & 3    & \cellcolor{gray!50}5.00$\pm$0.00   & \cellcolor{gray!50}540.87$\pm$0.00 & \cellcolor{gray!50}5     & \cellcolor{gray!50}540.87 & 1 \\
			RCdp5001$\vert$25 & \cellcolor{gray!50}9     & \cellcolor{gray!50}994.18 & 302146  & \cellcolor{gray!50}9     & \cellcolor{gray!50}994.18 & 17  & \cellcolor{gray!50}9.00$\pm$0.00   & \cellcolor{gray!50}994.18$\pm$0.00 & \cellcolor{gray!50}9     & \cellcolor{gray!50}994.18 & 1 \\
			RCdp5004$\vert$50 & $14^a$    & $1961.53^a$ & 774570  & \cellcolor{gray!50}6     & \cellcolor{gray!50}725.59 & 21    & 6.00$\pm$0.00   & 733.21$\pm$0.00 & 6 & 733.21 & 9 \\
			RCdp5007$\vert$50 & $13^a$    & $1814.33^a$ & 1427129  & \cellcolor{gray!50}7     & \cellcolor{gray!50}809.72 & 20  & \cellcolor{gray!50}7.00$\pm$0.00   & \cellcolor{gray!50}809.72$\pm$0.00 & \cellcolor{gray!50}7     & \cellcolor{gray!50}809.72 & 9 \\
			\bottomrule
	\end{tabular}}
	\label{tab:small_scale_results}
\end{table*}

We considered four state-of-the-art algorithms\footnote{The well-known TSP solver LKH (version 3.0) \cite{helsgaun2017extension} is also applicable to VRPSPDTW. However, in our preliminary experiments, we found LKH generally took prohibitively long time (typically two days) to find a feasible solution to the problem instances considered.
Therefore we did not consider it during our experiments.} 
in the comparative study, including Co-GA \cite{WangC12}, p-SA \cite{WangMZS15}, VNS-BSTS \cite{SHI2020103901}, and ALNS-PR\cite{HofS19}.
Each of these algorithms has been tested on part or all of the instances in the \textit{wc} set, and for each instance in the \textit{wc} set, at least one of the four algorithms has been reported to obtain the best-known solution.
Hence, for the \textit{wc} set, we directly obtained the best testing results of these algorithms from the original publications. 
For the new \textit{jd} set, we chose Co-GA as the competitor since it is the only open-sourced algorithm among the four, and the parameter settings reported in the original publication were used.

The detailed settings of MATE are summarized in Table~\ref{tab:alg_setting}.
MATE has four parameters that need to be set, i.e., $N$, $G_{max}$, $\omega_1$ and $\omega_2$, where $N$ is the population size, $G_{max}$ is the longest consecutive generations without improvement, and $\omega_1,\omega_2$ are the lower and upper bounds for the proportion of the nodes removed by the removal-and-reinsertion operator in the local search procedure (see Algorithm~\ref{alg:local_search}).
We set $G_{max}$ to 50 and set $\omega_1$ and $\omega_2$ to considerably large values, i.e., 0.2 and 0.4, respectively, such that the newly generated solution is likely to be very different from the original one.
For parameter $N$, given limited time budgets, there exists a trade-off between promoting the exploration in larger search space (i.e., large $N$) and facilitating more refined search in local areas (i.e., small $N$).
We consider the latter has direct and decisive impact on the algorithm's performance, especially when the search space is too large to sufficiently explore.
Therefore, for instances with $M \leq 100$, we set $N$ to 16.
Otherwise, we set $N$ to 36 (note $N$ must be a square number).
The effect of different values of $N$ will be further analyzed in Section~\ref{sec:sensitivity_N}.

All the experiments went through 30 independent runs, on an Intel Xeon E5-2699A v4 machine with 128 GB RAM and 22
cores (2.40 GHz, 55 MB Cache), running Centos 7.5.
Both MATE and Co-GA were implemented in C++.

\begin{table*}[tbp]
	\centering
	\caption{Comparison between Co-GA, p-SA, VNS-BSTS, ALNS-PR and the proposed MATE on medium-scale instances in the \textit{wc} set.
	    For each instance, the best performance regarding TC (i.e., $2000 \cdot \text{NV}+\text{TD}$) is highlighted in grey.
	    ``w-d-l'' refers to the number of win-draw-lose of the best results of MATE versus the best results of other algorithms.
	    ``No.best'' is the number of instances on which the algorithm finds the best solution.
	    ``Avg.Time'' refers to the averaged rescaled computation time.
	    ``$^*$'' indicates new best known solutions (i.e., never found before).
        ``N/A'' means not applicable.}
	\label{tab:medium_scale_results}
	\scalebox{0.82}{
	\begin{tabular}{lllllllllllllll}
		\toprule
		Instance & \multicolumn{2}{l}{Co-GA} & \multicolumn{2}{l}{p-SA} & \multicolumn{2}{l}{VNS-BSTS} & \multicolumn{3}{l}{ALNS-PR} & \multicolumn{5}{l}{MATE} \\
		& \multicolumn{2}{l}{} & \multicolumn{2}{l}{} & \multicolumn{2}{l}{} & \multicolumn{3}{l}{} & \multicolumn{2}{l}{Avg$\pm$std} & \multicolumn{2}{l}{Best} & \multicolumn{1}{l}{} \\
		\cmidrule(lr){2-3}
		\cmidrule(lr){4-5}
		\cmidrule(lr){6-7}
		\cmidrule(lr){8-10}
		\cmidrule(lr){11-12}
		\cmidrule(lr){13-14}
		\cmidrule(lr){15-15}
		& NV    & TD    & NV    & TD    & NV    & TD    & NV    & TD & Time   & NV    & TD    & NV    & TD & Time\\
		\midrule
		Rdp101 & 19    & 1653.53 & 19    & 1660.98 & \cellcolor{gray!50}19    & \cellcolor{gray!50}1650.80 & \cellcolor{gray!50}19    & \cellcolor{gray!50}1650.80 & 50.08 & \cellcolor{gray!50}19.00$\pm$0.00    & \cellcolor{gray!50}1650.80$\pm$0.00 & \cellcolor{gray!50}19    & \cellcolor{gray!50}1650.80 & 53.82 \\
		Rdp102 & 17    & 1488.04 & 17    & 1491.75 & \cellcolor{gray!50}17    & \cellcolor{gray!50}1486.12 & \cellcolor{gray!50}17    & \cellcolor{gray!50}1486.12 & 45.77 & \cellcolor{gray!50}17.00$\pm$0.00    & \cellcolor{gray!50}1486.12$\pm$0.00 & \cellcolor{gray!50}17    & \cellcolor{gray!50}1486.12 & 49.49\\
		Rdp103 & 14    & 1216.16 & 14    & 1226.77 & 13    & 1294.75 & 13    & 1297.01 & 39.06 & \cellcolor{gray!50}13.00$\pm$0.00    & \cellcolor{gray!50}1294.64$\pm$0.00 & \cellcolor{gray!50}13*    & \cellcolor{gray!50}1294.64* & 78.82 \\
		Rdp104 & 10    & 1015.41 & 10    & 1000.65 & \cellcolor{gray!50}10    & \cellcolor{gray!50}984.81 & \cellcolor{gray!50}10    & \cellcolor{gray!50}984.81 & 62.22 & \cellcolor{gray!50}10.00$\pm$0.00    & \cellcolor{gray!50}984.81$\pm$0.00 & \cellcolor{gray!50}10    & \cellcolor{gray!50}984.81 & 79.19\\
		Rdp105 & 15    & 1375.31 & 14    & 1399.81 & \cellcolor{gray!50}14    & \cellcolor{gray!50}1377.11 & \cellcolor{gray!50}14    & \cellcolor{gray!50}1377.11 & 38.52 & \cellcolor{gray!50}14.00$\pm$0.00    & \cellcolor{gray!50}1377.11$\pm$0.00 & \cellcolor{gray!50}14    & \cellcolor{gray!50}1377.11 & 105.43\\
		Rdp106 & 13    & 1255.48 & 12    & 1275.69 & 12    & 1261.40 & \cellcolor{gray!50}12    & \cellcolor{gray!50}1252.03 & 60.34 & \cellcolor{gray!50}12.00$\pm$0.00    & \cellcolor{gray!50}1252.03$\pm$0.00 & \cellcolor{gray!50}12    & \cellcolor{gray!50}1252.03 & 78.81\\
		Rdp107 & 11    & 1087.95 & 11    & 1082.92 & 10    & 1144.02 & \cellcolor{gray!50}10    & \cellcolor{gray!50}1121.86 & 41.17 & 10.00$\pm$0.00    & 1129.46$\pm$2.01 & 10    & 1124.90 & 78.99\\
		Rdp108 & 10    & 967.49 & 10    & 962.48 & 9     & 968.32 & 9     & 965.54 & 45.13 & 9.00$\pm$0.00   & 973.02$\pm$3.22 & \cellcolor{gray!50}9*     & \cellcolor{gray!50}965.22* & 79.71\\
		Rdp109 & 12    & 1160.00 & 12    & 1181.92 & 11    & 1224.86 & \cellcolor{gray!50}11    & \cellcolor{gray!50}1194.73 & 35.23 & 11.93$\pm$0.25 & 1161.71$\pm$18.43 & 11    & 1203.97 & 76.69\\
		Rdp110 & 12    & 1116.99 & 11    & 1106.52 & 11    & 1101.33 & \cellcolor{gray!50}10    & \cellcolor{gray!50}1148.20 & 41.91 & 10.97$\pm$0.18 & 1084.39$\pm$15.24 & 10    & 1166.47 & 78.42\\
		Rdp111 & 11    & 1065.27 & 11    & 1073.62 & 10    & 1117.76 & \cellcolor{gray!50}10    & \cellcolor{gray!50}1098.84 & 48.53 & \cellcolor{gray!50}10.00$\pm$0.00    & \cellcolor{gray!50}1098.84$\pm$0.00 & \cellcolor{gray!50}10    & \cellcolor{gray!50}1098.84 & 79.07\\
		Rdp112 & 10    & 974.03 & 10    & 966.06 & 10    & 961.29 & \cellcolor{gray!50}9     & \cellcolor{gray!50}1010.42 & 62.73 & 10.00$\pm$0.00    & 956.41$\pm$1.86 & 10    & 953.63 & 78.32\\
		Cdp101 & 11    & 1001.97 & 11    & 992.88 & \cellcolor{gray!50}11    & \cellcolor{gray!50}976.04 & \cellcolor{gray!50}11    & \cellcolor{gray!50}976.04 & 42.29 & \cellcolor{gray!50}11.00$\pm$0.00    & \cellcolor{gray!50}976.04$\pm$0.00 & \cellcolor{gray!50}11    & \cellcolor{gray!50}976.04 & 102.04\\
		Cdp102 & 10    & 961.38 & 10    & 955.31 & 10    & 942.45 & \cellcolor{gray!50}10    & \cellcolor{gray!50}941.49 & 61.59 & \cellcolor{gray!50}10.00$\pm$0.00    & \cellcolor{gray!50}941.49$\pm$0.00 & \cellcolor{gray!50}10    & \cellcolor{gray!50}941.49 & 78.18\\
		Cdp103 & 10    & 897.65 & 10    & 958.66 & 10    & 896.28 & \cellcolor{gray!50}10    & \cellcolor{gray!50}892.98 & 105.23 & 10.00$\pm$0.00    & 895.14$\pm$0.58 & \cellcolor{gray!50}10    & \cellcolor{gray!50}892.98 & 78.66\\
		Cdp104 & 10    & 878.93 & 10    & 944.73 & 10    & 872.39 & \cellcolor{gray!50}10    & \cellcolor{gray!50}871.40 & 101.90 & \cellcolor{gray!50}10.00$\pm$0.00   & \cellcolor{gray!50}871.40$\pm$0.00 & \cellcolor{gray!50}10    & \cellcolor{gray!50}871.40 & 79.41\\
		Cdp105 & 11    & 983.10 & 11    & 989.86 & 10    & 1080.63 & \cellcolor{gray!50}10    & \cellcolor{gray!50}1053.12 & 34.99 & 10.00$\pm$0.00    & 1074.51$\pm$0.00 & 10    & 1074.51 & 67.36\\
		Cdp106 & 11    & 878.29 & 11    & 878.29 & \cellcolor{gray!50}10    & \cellcolor{gray!50}963.45 & 10    & 967.71 & 38.03 & \cellcolor{gray!50}10.00$\pm$0.00    & \cellcolor{gray!50}963.45$\pm$0.00 & \cellcolor{gray!50}10    & \cellcolor{gray!50}963.45 & 100.87\\
		Cdp107 & 11    & 913.81 & 11    & 911.90 & \cellcolor{gray!50}10    & \cellcolor{gray!50}987.64 & \cellcolor{gray!50}10    & \cellcolor{gray!50}987.64 & 39.57 & 10.43$\pm$0.50 & 960.99$\pm$47.86 & 10    & 988.60 & 88.37\\
		Cdp108 & 10    & 951.24 & 10    & 1063.73 & 10    & 934.41 & 10    & 932.88 & 39.92 & 10.00$\pm$0.00    & 932.55$\pm$0.13 & \cellcolor{gray!50}10*    & \cellcolor{gray!50}932.49* & 76.52\\
		Cdp109 & 10    & 940.49 & 10    & 947.90 & \cellcolor{gray!50}10    & \cellcolor{gray!50}909.27 & 10    & 910.95 & 84.24 & \cellcolor{gray!50}10.00$\pm$0.00    & \cellcolor{gray!50}909.27$\pm$0.00 & \cellcolor{gray!50}10    & \cellcolor{gray!50}909.27 & 78.86\\
		RCdp101 & 15    & 1652.90 & 15    & 1659.59 & \cellcolor{gray!50}14    & \cellcolor{gray!50}1708.21 & 14    & 1776.58 & 23.20 & \cellcolor{gray!50}14.00$\pm$0.00    & \cellcolor{gray!50}1708.21$\pm$0.00 & \cellcolor{gray!50}14    & \cellcolor{gray!50}1708.21 & 27.13\\
		RCdp102 & 14    & 1497.05 & 13    & 1522.76 & 13    & 1526.36 & 12    & 1583.62 & 42.04 & 12.00$\pm$0.00    & 1586.62$\pm$7.31 & \cellcolor{gray!50}12*    & \cellcolor{gray!50}1570.28* & 78.92\\
		RCdp103 & 12    & 1338.76 & 11    & 1344.62 & 11    & 1336.05 & 11    & 1283.52 & 63.71 & 11.00$\pm$0.00    & 1284.66$\pm$0.85 & \cellcolor{gray!50}11*    & \cellcolor{gray!50}1282.53* & 78.64\\
		RCdp104 & 11    & 1188.49 & 10    & 1268.43 & 10    & 1177.21 & 10    & 1171.65 & 49.45 & 10.00$\pm$0.00    & 1172.37$\pm$1.06 & \cellcolor{gray!50}10*    & \cellcolor{gray!50}1171.37* & 79.88\\
		RCdp105 & 14    & 1581.26 & 14    & 1581.54 & 14    & 1548.38 & 14    & 1548.96 & 37.14 & 13.87$\pm$0.34 & 1566.46$\pm$46.91 & \cellcolor{gray!50}13*    & \cellcolor{gray!50}1646.36* & 72.62\\
		RCdp106 & 13    & 1422.87 & 13    & 1418.16 & 12    & 1408.19 & \cellcolor{gray!50}12    & \cellcolor{gray!50}1392.47 & 44.85 & \cellcolor{gray!50}12.00$\pm$0.00    & \cellcolor{gray!50}1392.47$\pm$0.00 & \cellcolor{gray!50}12    & \cellcolor{gray!50}1392.47 & 78.62\\
		RCdp107 & 12    & 1282.10 & 11    & 1360.17 & 11    & 1295.43 & 11    & 1255.06 & 46.49 &\cellcolor{gray!50}11.00$\pm$0.00    & \cellcolor{gray!50}1252.79$\pm$0.00 & \cellcolor{gray!50}11*    & \cellcolor{gray!50}1252.79* & 78.67\\
		RCdp108 & 11    & 1175.04 & 11    & 1169.57 & 10    & 1207.60 & \cellcolor{gray!50}10    & \cellcolor{gray!50}1198.36 & 42.77 & 10.83$\pm$0.37 & 1163.77$\pm$27.35 & 10    & 1208.58 & 79.18\\
		Rdp201 & 4     & 1280.44 & 4     & 1286.55 & 4     & 1254.57 & 4     & 1253.23 & 73.26 & 4.00$\pm$0.00     & 1252.55$\pm$0.29 & \cellcolor{gray!50}4*     & \cellcolor{gray!50}1252.37* & 78.80\\
		Rdp202 & 4     & 1100.92 & 4     & 1150.31 & 3     & 1202.27 & \cellcolor{gray!50}3     & \cellcolor{gray!50}1191.70 & 102.34 & 3.70$\pm$0.46 & 1141.45$\pm$94.90 & 3     & 1223.69 & 77.46\\
		Rdp203 & 3     & 950.79 & 3     & 997.84 & 3     & 949.42 & 3     & 946.28 & 186.03 & 3.00$\pm$0.00     & 946.18$\pm$1.32 & \cellcolor{gray!50}3*     & \cellcolor{gray!50}939.58* & 73.94\\
		Rdp204 & 3     & 775.23 & 2     & 848.01 & 2     & 837.13 & \cellcolor{gray!50}2     & \cellcolor{gray!50}833.09 & 244.33 & 2.70$\pm$0.46 & 789.85$\pm$62.60 & 2     & 835.28 & 78.54\\
		Rdp205 & 3     & 1064.43 & 3     & 1046.06 & 3     & 1027.49 & \cellcolor{gray!50}3     & \cellcolor{gray!50}994.43 & 176.04 &3.00$\pm$0.00     & 997.83$\pm$2.55 & \cellcolor{gray!50}3     & \cellcolor{gray!50}994.43 & 80.49\\
		Rdp206 & 3     & 961.32 & 3     & 959.94 & 3     & 938.63 & 3     & 913.68 & 196.92 & 3.00$\pm$0.00     & 908.69$\pm$2.10 & \cellcolor{gray!50}3*     & \cellcolor{gray!50}906.14* & 78.23\\
		Rdp207 & 3     & 835.01 & 2     & 899.82 & 2     & 912.26 & \cellcolor{gray!50}2     & \cellcolor{gray!50}890.61 & 180.77 & 3.00$\pm$0.00     & 814.90$\pm$0.63 & 3     & 811.51 & 71.13\\
		Rdp208 & 3     & 718.51 & 2     & 739.06 & 2     & 737.26 & \cellcolor{gray!50}2     & \cellcolor{gray!50}726.82 & 219.64 & 2.00$\pm$0.00     & 733.41$\pm$2.44 & \cellcolor{gray!50}2     & \cellcolor{gray!50}726.82 & 81.23\\
		Rdp209 & 3     & 930.26 & 3     & 947.80 & 3     & 940.29 & \cellcolor{gray!50}3     & \cellcolor{gray!50}909.16 & 189.71 & 3.00$\pm$0.00     & 917.77$\pm$1.82 & \cellcolor{gray!50}3     & \cellcolor{gray!50}909.16 & 76.27\\
		Rdp210 & 3     & 983.75 & 3     & 1005.11 & 3     & 945.97 & \cellcolor{gray!50}3     & \cellcolor{gray!50}939.37 & 181.50 & 3.00$\pm$0.00     & 950.44$\pm$3.69 & \cellcolor{gray!50}3     & \cellcolor{gray!50}939.37 & 78.07\\
		Rdp211 & 3     & 839.61 & 3     & 812.44 & 3     & 805.22 & \cellcolor{gray!50}2     & \cellcolor{gray!50}904.44 & 188.14 & 3.00$\pm$0.00     & 778.62$\pm$1.50 & 3     & 767.82 & 68.64\\
		Cdp201 & \cellcolor{gray!50}3     & \cellcolor{gray!50}591.56 & \cellcolor{gray!50}3     & \cellcolor{gray!50}591.56 & \cellcolor{gray!50}3     & \cellcolor{gray!50}591.56 & \cellcolor{gray!50}3     & \cellcolor{gray!50}591.56 & 53.39 & \cellcolor{gray!50}3.00$\pm$0.00     & \cellcolor{gray!50}591.56$\pm$0.00 & \cellcolor{gray!50}3     & \cellcolor{gray!50}591.56 & 90.21\\
		Cdp202 & \cellcolor{gray!50}3     & \cellcolor{gray!50}591.56 & \cellcolor{gray!50}3     & \cellcolor{gray!50}591.56 & \cellcolor{gray!50}3     & \cellcolor{gray!50}591.56 & \cellcolor{gray!50}3     & \cellcolor{gray!50}591.56 & 100.98 & \cellcolor{gray!50}3.00$\pm$0.00     & \cellcolor{gray!50}591.56$\pm$0.00 & \cellcolor{gray!50}3     & \cellcolor{gray!50}591.56 & 158.59\\
		Cdp203 & \cellcolor{gray!50}3     & \cellcolor{gray!50}591.17 & \cellcolor{gray!50}3     & \cellcolor{gray!50}591.17 & \cellcolor{gray!50}3     & \cellcolor{gray!50}591.17 & \cellcolor{gray!50}3     & \cellcolor{gray!50}591.17 & 96.82 & \cellcolor{gray!50}3.00$\pm$0.00     & \cellcolor{gray!50}591.17$\pm$0.00 & \cellcolor{gray!50}3     & \cellcolor{gray!50}591.17 & 62.22\\
		Cdp204 & \cellcolor{gray!50}3     & \cellcolor{gray!50}590.60 & 3     & 594.07 & 3     & 599.33 & \cellcolor{gray!50}3     & \cellcolor{gray!50}590.60 & 113.40 & \cellcolor{gray!50}3.00$\pm$0.00     & \cellcolor{gray!50}590.60$\pm$0.00 & \cellcolor{gray!50}3     & \cellcolor{gray!50}590.60 & 78.93\\
		Cdp205 & \cellcolor{gray!50}3     & \cellcolor{gray!50}588.88 & \cellcolor{gray!50}3     & \cellcolor{gray!50}588.88 & \cellcolor{gray!50}3     & \cellcolor{gray!50}588.88 & \cellcolor{gray!50}3     & \cellcolor{gray!50}588.88 & 79.64 & \cellcolor{gray!50}3.00$\pm$0.00     & \cellcolor{gray!50}588.88$\pm$0.00 & \cellcolor{gray!50}3     & \cellcolor{gray!50}588.88 & 134.77\\
		Cdp206 & \cellcolor{gray!50}3     & \cellcolor{gray!50}588.49 & \cellcolor{gray!50}3     & \cellcolor{gray!50}588.49 & \cellcolor{gray!50}3     & \cellcolor{gray!50}588.49 & \cellcolor{gray!50}3     & \cellcolor{gray!50}588.49 & 79.73 & \cellcolor{gray!50}3.00$\pm$0.00     & \cellcolor{gray!50}588.49$\pm$0.00 & \cellcolor{gray!50}3     & \cellcolor{gray!50}588.49 & 176.76\\
		Cdp207 & \cellcolor{gray!50}3     & \cellcolor{gray!50}588.29 & \cellcolor{gray!50}3     & \cellcolor{gray!50}588.29 & \cellcolor{gray!50}3     & \cellcolor{gray!50}588.29 & \cellcolor{gray!50}3     & \cellcolor{gray!50}588.29 & 87.26 & \cellcolor{gray!50}3.00$\pm$0.00     & \cellcolor{gray!50}588.29$\pm$0.00 & \cellcolor{gray!50}3     & \cellcolor{gray!50}588.29 & 196.92\\
		Cdp208 & \cellcolor{gray!50}3     & \cellcolor{gray!50}588.32 & 3     & 599.32 & \cellcolor{gray!50}3     & \cellcolor{gray!50}588.32 & \cellcolor{gray!50}3     & \cellcolor{gray!50}588.32 & 75.52 & \cellcolor{gray!50}3.00$\pm$0.00     & \cellcolor{gray!50}588.32$\pm$0.00 & \cellcolor{gray!50}3     & \cellcolor{gray!50}588.32 & 215.96\\
		RCdp201 & 4     & 1587.92 & 4     & 1513.72 & 4     & 1437.48 & \cellcolor{gray!50}4     & \cellcolor{gray!50}1406.94 & 58.45 & 4.00$\pm$0.00     & 1407.16$\pm$0.36 & \cellcolor{gray!50}4     & \cellcolor{gray!50}1406.94 & 78.89\\
		RCdp202 & 4     & 1211.12 & 4     & 1273.26 & \cellcolor{gray!50}3     & \cellcolor{gray!50}1412.52 & 3     & 1414.55 & 87.51 & 4.00$\pm$0.00     & 1161.29$\pm$0.00 & 4     & 1161.29 & 79.10\\
		RCdp203 & 4     & 964.65 & 3     & 1123.58 & 3     & 1064.95 & \cellcolor{gray!50}3     & \cellcolor{gray!50}1050.64 & 183.93 & 3.00$\pm$0.00     & 1071.13$\pm$4.05 & 3     & 1056.96 & 77.17\\
		RCdp204 & 3     & 822.02 & 3     & 897.14 & 3     & 813.74 & \cellcolor{gray!50}3     & \cellcolor{gray!50}798.46 & 207.81 & 3.00$\pm$0.00     & 798.82$\pm$0.21 & \cellcolor{gray!50}3     & \cellcolor{gray!50}798.46 & 71.01\\
		RCdp205 & 4     & 1410.18 & 4     & 1371.08 & 4     & 1316.06 & \cellcolor{gray!50}4     & \cellcolor{gray!50}1297.65 & 73.57 & \cellcolor{gray!50}4.00$\pm$0.00     & \cellcolor{gray!50}1297.65$\pm$0.00 & \cellcolor{gray!50}4     & \cellcolor{gray!50}1297.65 & 77.82\\
		RCdp206 & 3     & 1176.85 & 3     & 1166.88 & 3     & 1154.36 & \cellcolor{gray!50}3     & \cellcolor{gray!50}1146.32 & 130.05 & 3.00$\pm$0.00     & 1156.23$\pm$2.96 & \cellcolor{gray!50}3     & \cellcolor{gray!50}1146.32 & 78.54\\
		RCdp207 & 4     & 1036.59 & 3     & 1089.85 & 3     & 1098.64 & 3     & 1061.84 & 173.22 & 3.00$\pm$0.00     & 1073.42$\pm$1.94 & \cellcolor{gray!50}3*     & \cellcolor{gray!50}1061.14* & 84.47\\
		RCdp208 & 3     & 878.57 & 3     & 862.89 & 3     & 843.30 & \cellcolor{gray!50}3     & \cellcolor{gray!50}828.14 & 159.46 & 3.00$\pm$0.00     & 833.40$\pm$1.11 & 3     & 828.44 & 72.98\\
		\midrule
		w-d-l & \multicolumn{2}{l}{48-8-0} & \multicolumn{2}{l}{49-6-1} & \multicolumn{2}{l}{36-15-5} & \multicolumn{3}{l}{15-27-14} & \multicolumn{2}{l}{N/A} & \multicolumn{3}{l}{N/A} \\
		No.best & \multicolumn{2}{l}{8} & \multicolumn{2}{l}{6} & \multicolumn{2}{l}{17} & \multicolumn{3}{l}{40} & \multicolumn{2}{l}{24} & \multicolumn{3}{l}{42} \\
		Avg.Time & \multicolumn{2}{l}{} & \multicolumn{2}{l}{} & \multicolumn{2}{l}{} & \multicolumn{2}{l}{} & \multicolumn{1}{l}{92.28} &\multicolumn{2}{l}{} & \multicolumn{2}{l}{} & \multicolumn{1}{l}{86.03}\\
		\bottomrule
	\end{tabular}}
\end{table*}

\subsection{Results on Small-scale and Medium-scale Instances}
\label{sec:small_medium_analysis}
The commercial mathematical programming software CPLEX has been used in \cite{WangC12} to find the optimal solutions for the small-scale instances ($M \leq 50$) in the \textit{wc} set.
The best results of CPLEX and Co-GA on these instances, obtained from \cite{WangC12}, in comparison with the testing results of MATE, are presented in Table~\ref{tab:small_scale_results}.
Moreover, to make the comparison as fair as possible, we rescale the computation time of all approaches into a common time measure that takes into account the used CPUs.
Specifically, we relate the Passmark scores of the CPUs used in the computational studies of the papers to the score of our E5-2699 V4.
Each Passmark score refers to the performance of a single core of the respective CPU\footnote{The Passmark score of the Core2 Quad 2.4G used in \cite{WangC12} is 957. For the i5-6600 used in \cite{HofS19}, it is 2272. For our E5-2699 V4, it is 1037. See \url{www.passmark.com} for more details.}.
The rescaled times in seconds are given in Table~\ref{tab:small_scale_results}.
Note that for the \textit{wc} set, minimizing NV has a higher priority than minimizing TD; this is achieved by setting $u_1$ and $u_2$ in Eq.~(\ref{eq:vrpspdtw}) to 2000 and 1, respectively.
Therefore for these instances, the TC of a solution equals to $2000 \cdot \text{NV} + \text{TD}$, and a solution is better than another if the former has smaller TC.
From Table~\ref{tab:small_scale_results} it can be observed that CPLEX managed to solve five of the instances, on which MATE and Co-GA also found optimal solutions.
For the other four instances, CPLEX prematurely terminated due to the ``out-of-memory'' condition, while the two meta-heuristic algorithms found much better solutions than CPLEX, in a much shorter time.
Moreover, MATE consumed less computation time than Co-GA, although they achieved very close solution quality.
Overall, both of them performed much better than CPLEX, with the performance gap becoming  even bigger as the problem scale increasing.
It is worth mentioning that MATE performed very stably on these instances, achieving the standard deviation of 0 across 30 independent runs.

\begin{table*}[tbp]
	\centering
	\caption{Comparison between Co-GA and the proposed MATE on large-scale instances in the $jd$ set. The total costs (TC) are reported.
		For each instance, the best performance is highlighted in grey; the average performance of an algorithm is indicated in bold if it is significantly better than the other algorithm based on 30 independent runs, according to the Wilcoxon rank-sum test with significance level p = 0.05.}
	\label{tab:large_scale_results}
	\scalebox{0.9}{
		\begin{tabular}{llllllll}
			\toprule
			Instance$\vert M$ & \multicolumn{3}{l}{Co-GA} & \multicolumn{3}{l}{MATE} & p-value \\
			\cmidrule(lr){2-4}
			\cmidrule(lr){5-7}
			& Avg$\pm$std & Best & Time  & Avg$\pm$std & Best & Time & \\
			\midrule
			F201$\vert$200 & 67506$\pm$712 & 69301 & 675 & \textbf{66097}$\pm$292 & \cellcolor{gray!50}65106 & 133 & 0.0000 \\
			F202$\vert$200 & 66734$\pm$766 & 68513 &  2423 & \textbf{66038}$\pm$422 & \cellcolor{gray!50}65012 & 202 & 0.0002 \\
			F203$\vert$200 & 67651$\pm$713 & 69480 & 1046  & \textbf{67090}$\pm$332 & \cellcolor{gray!50}65980  & 240 &0.0010 \\
			F204$\vert$200 & 66247$\pm$590 & 67557 & 773  & \textbf{65851}$\pm$326 & \cellcolor{gray!50}64747 & 180 & 0.0071 \\
			F401$\vert$400 & 127626$\pm$960 & 129258 & 7200  & \textbf{123261}$\pm$446 & \cellcolor{gray!50}122319 & 773 & 0.0000  \\
			F402$\vert$400 & 133307$\pm$1069 & 135032 & 7200  & \textbf{128091}$\pm$410 & \cellcolor{gray!50}126887 & 341 & 0.0000  \\
			F403$\vert$400 & 126660$\pm$1212 & 129191 & 7200  & \textbf{122306}$\pm$682 & \cellcolor{gray!50}120130 & 656 & 0.0000  \\
			F404$\vert$400 & 130807$\pm$993 & 133203 & 6026  & \textbf{125242}$\pm$359 & \cellcolor{gray!50}124517  & 1995 & 0.0000  \\
			F601$\vert$600 & 195250$\pm$1599 & 198688 & 7200  & \textbf{184119}$\pm$608 & \cellcolor{gray!50}182504 & 1161 & 0.0000  \\
			F602$\vert$600 & 202907$\pm$1235 & 205101 & 7200  & \textbf{188891}$\pm$645 & \cellcolor{gray!50}187236 & 1767 & 0.0000  \\
			F603$\vert$600 & 201579$\pm$1809 & 205827 & 7200 & \textbf{188050}$\pm$621 & \cellcolor{gray!50}186644  & 914 & 0.0000  \\
			F604$\vert$600 & 200232$\pm$1830 & 203499 & 7200  & \textbf{188110}$\pm$790 & \cellcolor{gray!50}186289  & 2465 & 0.0000  \\
			F801$\vert$800 & 235445$\pm$2870 & 241995 & 7200  & \textbf{214634}$\pm$561 & \cellcolor{gray!50}213661  & 1506 & 0.0000  \\
			F802$\vert$800 & 238457$\pm$1500 & 240912 & 7200  & \textbf{213276}$\pm$292 & \cellcolor{gray!50}212752  & 2560 & 0.0000  \\
			F803$\vert$800 & 236211$\pm$2136 & 241150 & 7200  & \textbf{214870}$\pm$318 & \cellcolor{gray!50}214126  & 2288 & 0.0000  \\
			F804$\vert$800 & 231765$\pm$1959 & 237389 & 7200  & \textbf{210845}$\pm$429 & \cellcolor{gray!50}209431  & 3497 & 0.0000  \\
			F1001$\vert$1000 & N/A & N/A & N/A & \textbf{314914}$\pm$1096 & \cellcolor{gray!50}312606 & 2014 &  N/A\\
			F1002$\vert$1000 & N/A & N/A & N/A & \textbf{311718}$\pm$1153 & \cellcolor{gray!50}309158 & 5100 &  N/A\\
			F1003$\vert$1000 & N/A & N/A & N/A & \textbf{313989}$\pm$981 & \cellcolor{gray!50}311377  & 4403 &  N/A\\
			F1004$\vert$1000 & N/A & N/A & N/A & \textbf{311415}$\pm$943 & \cellcolor{gray!50}308816  & 3712 &  N/A\\
			\bottomrule
	\end{tabular}}
\end{table*}

In general, the algorithms' performamce on instances of larger scales is of more interest.
Table~\ref{tab:medium_scale_results} presents the testing results of MATE and the best results of the four compared algorithms, obtained from the original publications, on the medium-scale instances ($M = 100$) in the \textit{wc} set.
The rescaled computation times of MATE and the best-performing algorithm among the competitors (i.e., ALNS-PR), are also presented.
The efficacy of MATE can be evaluated from two perspectives, i.e., the best and the average performance it has achieved in 30 independent runs.
We first take a closer look at the best performance since the compared algorithms' results reported in Table~\ref{tab:medium_scale_results} are their best performance.
Overall, MATE is the best-performing algorithm in Table~\ref{tab:medium_scale_results}.
From the row headed ``No.best'', it can be found that MATE obtained the best solutions on 42 out of 56 instances, which is the most among all algorithms.
More specifically, these instances are evenly distributed across all instance subsets, including 8 out of 12 in the Rdp1{\raise.17ex\hbox{$\scriptstyle\sim$}} subset, 7 out of 9 in the Cdp1{\raise.17ex\hbox{$\scriptstyle\sim$}} subset, 7 out of 8 in the RCdp1{\raise.17ex\hbox{$\scriptstyle\sim$}} subset, 7 out of 11 in the Rdp2{\raise.17ex\hbox{$\scriptstyle\sim$}} subset, 8 out of 8 in the Cdp2{\raise.17ex\hbox{$\scriptstyle\sim$}} subset, and 5 out of 8 in the RCdp2{\raise.17ex\hbox{$\scriptstyle\sim$}} subset.
Considering the node distribution and constraint intensity are different across different instance types (see Section~\ref{sec:benchmark}), such results indicate that MATE performs both strongly and robustly.
In comparison, ALNS-PR, the second best-performing algorithm, performed not well on the RCdp1{\raise.17ex\hbox{$\scriptstyle\sim$}} subset, finding best solutions for only 2 out of 8 instances in the set.

\textcolor{black}
{
The row headed ``w-d-l'' in Table~\ref{tab:medium_scale_results} presents the statistics of comparing the best performance of MATE with the best performance of other algorithms.
Compared to Co-GA, p-SA and VNS-BSTS, MATE exhibits remarkable performance advantage.
This is due to the fact that, unlike MATE, these algorithms have rather limited ability in exploring the search space since they neither construct high-diversity initial solutions nor adopt large-step-size operators to escape from local optima.
As a result, MATE can usually identify more promising regions than others, and therefore have higher probability of finding high-quality solutions.
Although the performance advantage of MATE over ALNS-PR is slight, the former still shows higher efficiency and consumes less computation time than the latter.
This is due to the fact that MATE incorporate the constant-time-complexity move evaluation approach, which can dramatically improve the search efficiency.
In Section~\ref{sec:ablation_study} we will further verify the effectiveness of all the novel components integrated in MATE.
}
Finally, it is notable that MATE managed to find \textbf{new} best-known solutions on 12 instances in the \textit{wc} set, especially considering this benchmark set has been widely used in the literature for more than 9 years.


It is worth mentioning on 24 out of 56 instances, MATE consistently found the best solution in every single run across 30 independent runs, and on almost half of the instances, i.e., 26 out of 56, the average performance of MATE across 30 independent runs is the same as its best performance.
Comparing to the best performance of Co-GA, p-SA and VNS-BSTS, the average performance of MATE is still significantly better.
Such results demonstrate that MATE performs very stably, although it is a randomized algorithm in nature.
In conclusion, all of the above observations are evidence that MATE performs better than the state-of-the-art algorithms on a wide range of problem instances, and in particular, it has significantly raised the performance bar on the \textit{wc} set.

\subsection{Results on Large-scale Instances}
Table~\ref{tab:large_scale_results} presents the testing results of Co-GA and MATE on large-scale instances in the \textit{jd} set.
To prevent the algorithms from running prohibitively long, we limit the maximum computation time to 7200 seconds.
Unlike the \textit{wc} set, for this set there is no priority for minimizing either NV or TD; thus the TC of the solutions are reported.
In the experiments, Co-GA kept crashing on instances with 1000 customers; for these instances ``N/A'' is reported.
Moreover, nonparametric tests (Wilcoxon rank-sum test with significance level $p=0.05$) were conducted to compare the performance of Co-GA and MATE, with p-value shown in the last column.
It  can be seen that MATE obtained significantly better solutions than Co-GA on every instance in the \textit{jd} set, and also consumed much less computation time than Co-GA.
The gap between the best performance of Co-GA and MATE across 30 independent runs, i.e., $[TC_1-TC_2]/TC_2$ where $TC_1$ and $TC_2$ are the best performance of Co-GA and MATE respectively, is 5.37\%, 6.65\%, 9.48\% and 13.12\%, averaged on 200-customer, 400-customer, 600-customer and 800-customer instances, respectively.
For average performance, the corresponding gaps are 1.16\%, 3.90\%, 6.78\%, and 10.34\%.
It can be observed as the problem scale grows, the performance gap between Co-GA and MATE also becomes larger.

\begin{table}[tbp]
	\centering
	\caption{Average PDR for the five MATE variants on each subset of the \textit{wc} set.
		For each subset, the highest average PDR is highlighted in grey.
		``Avg.PDR'' refers to the average PDR on all instances in the \textit{wc} set.}
	\scalebox{0.9}{
		\begin{tabular}{llllll}
			\toprule
			Subset     & w/o ED & w/o RI & w/o LS & w/o ES & w/o FL\\
			\midrule
			$\textit{wc}_{small}$    & 0.00\% & 0.00\% & \cellcolor{gray!50}0.01\%  & 0.00\%  & 0.00\%\\
			Rdp1{\raise.17ex\hbox{$\scriptstyle\sim$}}   & 0.21\% & 0.18\% & \cellcolor{gray!50}1.17\% & 0.35\% & 0.78\% \\
			Cdp1{\raise.17ex\hbox{$\scriptstyle\sim$}}   & \cellcolor{gray!50}1.24\% & 1.07\% & 0.55\% & 0.76\% & 0.12\% \\
			RCdp1{\raise.17ex\hbox{$\scriptstyle\sim$}}   & 0.80\% & 0.21\% & \cellcolor{gray!50}1.53\% & 1.41\% & 0.46\% \\
			Rdp2{\raise.17ex\hbox{$\scriptstyle\sim$}}   & 1.14\% & 0.94\% & \cellcolor{gray!50}1.77\% & 0.64\% & 0.66\% \\
			Cdp2{\raise.17ex\hbox{$\scriptstyle\sim$}}  & 0.00\% & 0.00\% & \cellcolor{gray!50}0.37\% & 0.01\% & 0.09\%\\
			RCdp2{\raise.17ex\hbox{$\scriptstyle\sim$}}  & 0.03\% & 0.43\% & \cellcolor{gray!50}2.14\% & 0.17\% & 0.40\%\\
			\midrule
			Avg.PDR   &0.51\%  & 0.42\% & \cellcolor{gray!50}1.06\% & 0.47\% & 0.38\%\\
			\bottomrule
	\end{tabular}}
	\label{tab:ablation_study_results}
\end{table}

\subsection{Effectiveness of Each Component in MATE}
\label{sec:ablation_study}
An ablation study was conducted to further assess the effectiveness of the novel components integrated into MATE.
More specifically, we tested the following 5 different variants of MATE on the \textit{wc} set, each of which is different from MATE in one component:
\begin{enumerate}
  \item w/o ED: In initialization, it uses random values drawn from [0,1] for $(\lambda, \gamma)$, instead of the proposed evenly-distributed values (lines 4-5 of Algorithm~\ref{alg:initialization});
  \item w/o RI: After doing route inheritance during crossover, it adopts the two-step insertion as described in \cite{AlvarengaMT07} (see Section~\ref{sec:crossover}) to reinsert the unassigned customers, instead of the proposed regret-based insertion (lines 7-15 of Algorithm~\ref{alg:crossover});
  \item w/o LS: It has no local search procedure, i.e., line 8 in Algorithm~\ref{alg:mate} is removed;
  \item w/o ES: In the local search procedure, it does not seek to escape from local optimum with the removal-and-reinsertion operator, i.e., lines 2-12 are removed from Algorithm~\ref{alg:local_search};
  \item w/o FL: In the local search procedure, it does not seek to identify local optimum, i.e., line 1 and line 9 in Algorithm~\ref{alg:local_search} are removed.
\end{enumerate}
For each of the above variant, the performance degradation ratio (PDR) is computed as $[TC_1-TC_2]/TC_2$ for each instance, where $TC_1$ and $TC_2$ are the average TC obtained by the variant and MATE across 30 independent runs.
Therefore, the component being examined is useful for this specific instance only if the $\text{PDR} > 0$, and the larger it is, the greater the component contributes to the performance of MATE.
Table~\ref{tab:ablation_study_results} presents the PDR averaged on each instance subset of the \textit{wc} set, where $\textit{wc}_{small}$ refers to all the small-scale instances.
In can be observed that, on average, none of the examined components has negative effect on the performance of MATE, and in most cases, they make MATE perform better.
In particular, the results in the ``w/o ED'' column indicate that using evenly-distributed values for $(\lambda, \gamma)$ is better than using random values, since the former enables more sufficient exploration in the design space of $(\lambda, \gamma)$.
The results in the ``w/o RI'' column indicate the regret-based insertion performs better than the two-step insertion, which is appealing considering the former is even simpler.
As we expect, the results in the ``w/o LS'' column demonstrate that the local search procedure has the most significant contribution to the performance of MATE, and it is also the only one achieving positive effect on all subsets.
The results in the last two columns (``w/o ES'' and ``w/o FL'') indicate that removing either small-neighborhood operators (i.e., identifying local optimum) or large-neighborhood operator (i.e., escape from local optimum) from the local search procedure will result in performance degradation, which on the other hand indicate that the strategy of switching between operators of different step sizes is effective in obtaining high-quality solutions.

\begin{figure}[t]
	\centering
	\scalebox{1.0}{\includegraphics[width=\linewidth]{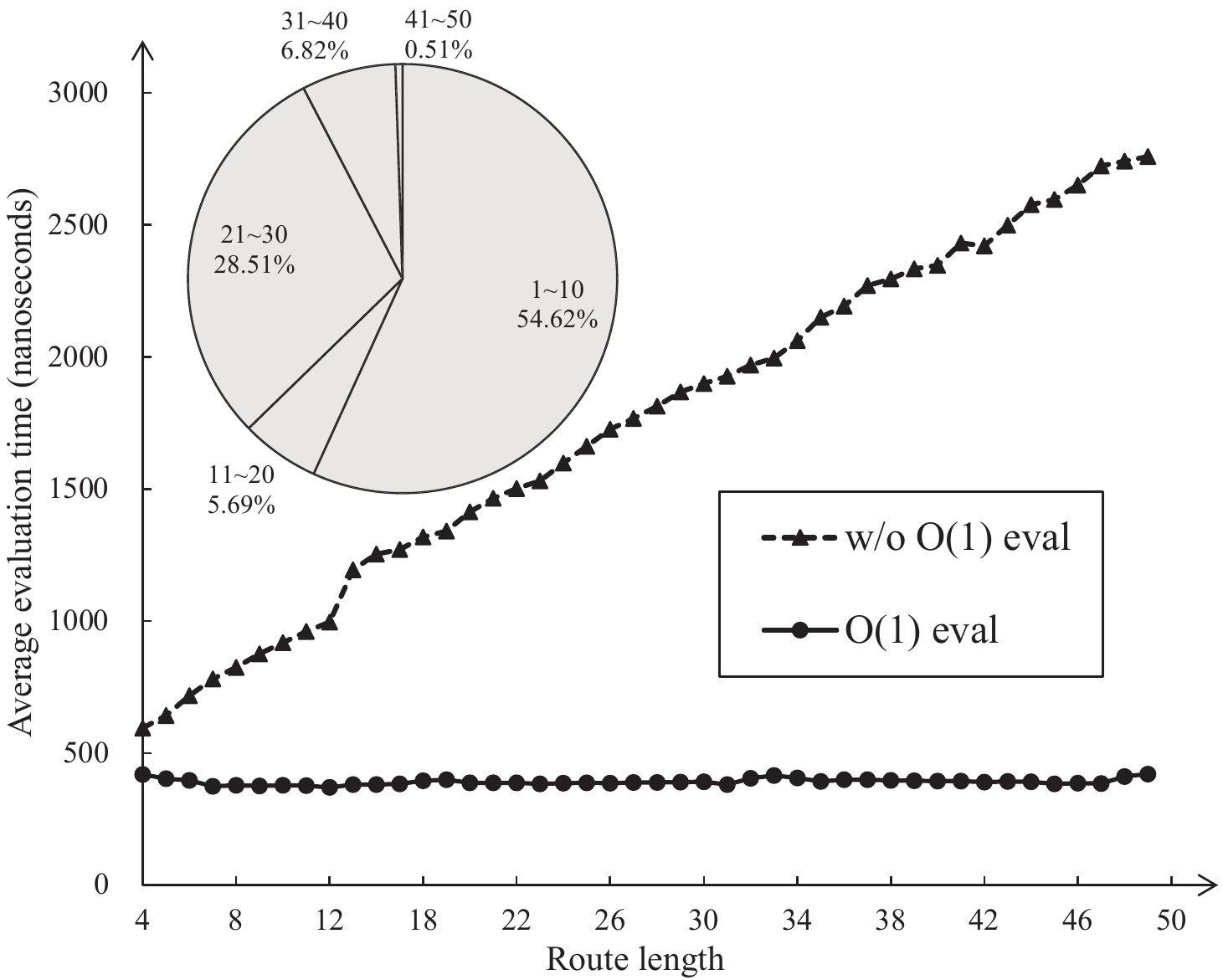}}
	\caption{Average evaluation time (ns) at different route lengths when turning on/off the new evaluation approach. 
		    The proportions of different route lengths are also illustrated.}
	\label{fig:results_O_1}
\end{figure}

\begin{table*}[tbp]
	\centering
	\caption{Average PDR for 7 different $N$ values on different subsets.
		For each subset, the lowest average PDR is highlighted in grey.
		``$M$'' is the number of customers of each instance in the subset.}
	\scalebox{1.0}{
		\begin{tabular}{llllllll}
			\toprule
			Subset$\vert$\textit{M} & \textit{N}=4  & \textit{N}=9  & \textit{N}=16  & \textit{N}=25  & \textit{N}=36  & \textit{N}=49  & \textit{N}=64 \\
			\midrule
			$\textit{wc}_{small}\vert$10,25,50 & \cellcolor{gray!50}0.00\% & \cellcolor{gray!50}0.00\% & \cellcolor{gray!50}0.00\% & \cellcolor{gray!50}0.00\% & \cellcolor{gray!50}0.00\% & \cellcolor{gray!50}0.00\% & \cellcolor{gray!50}0.00\% \\
			Rdp1{\raise.17ex\hbox{$\scriptstyle\sim$}}$\vert$100 & 0.67\% & 0.48\% & 0.90\% & 0.67\% & \cellcolor{gray!50}0.09\% & 1.16\% & 1.27\% \\
			Cdp1{\raise.17ex\hbox{$\scriptstyle\sim$}}$\vert$100 & 0.79\% & 1.37\% & 1.17\% & 1.38\% & 0.42\% &\cellcolor{gray!50}0.00\% & 1.38\% \\
			Rdp2{\raise.17ex\hbox{$\scriptstyle\sim$}}$\vert$100 & 1.19\% & 1.21\% & 1.17\% & 1.78\% & 0.28\% & 1.79\% & \cellcolor{gray!50}0.00\% \\
			Cdp2{\raise.17ex\hbox{$\scriptstyle\sim$}}$\vert$100 &\cellcolor{gray!50}0.00\% & \cellcolor{gray!50}0.00\% & \cellcolor{gray!50}0.00\% & \cellcolor{gray!50}0.00\% & \cellcolor{gray!50}0.00\% & \cellcolor{gray!50}0.00\% & \cellcolor{gray!50}0.00\% \\
			RCdp1{\raise.17ex\hbox{$\scriptstyle\sim$}}$\vert$100 & 1.01\% & 1.40\% & 0.92\% & 2.13\% & 0.95\% & \cellcolor{gray!50}0.09\% & 1.49\% \\
			RCdp2{\raise.17ex\hbox{$\scriptstyle\sim$}}$\vert$100 & 0.09\% & 0.09\% & 0.08\% & 0.10\% & \cellcolor{gray!50}0.00\% & 0.11\% & \cellcolor{gray!50}0.09\% \\
			F20{\raise.17ex\hbox{$\scriptstyle\sim$}}$\vert$200 &\cellcolor{gray!50}0.00\% & 0.56\% & 0.77\% & 1.88\% & 2.11\% & 2.48\% & 2.22\% \\
			F40{\raise.17ex\hbox{$\scriptstyle\sim$}}$\vert$400 & 0.42\% & 0.29\% & \cellcolor{gray!50}0.21\% & 0.68\% & 0.64\% & 0.57\% & 0.63\% \\
			\bottomrule
	\end{tabular}}
	\label{tab:analysis_N_results}
\end{table*}

One may notice that the new move evaluation approach was left out from the above analysis.
Since it does not affect any algorithmic behavior of MATE, but serves as an accelerator, in the experiments we kept tack of the average computation time spent on move evaluation at different route lengths, with the new approach turned on/off.
Note that when it was turned off, the traditional approach which would traverse the involved routes was used.
The results are presented in Figure~\ref{fig:results_O_1}, where ``O(1) eval'' refers to the new evaluation approach.
In addition, the proportions of different lengths of routes encountered through the whole experiments are also shown.
It can be clearly seen that the move evaluation time for the traditional approach grows linearly with route length, while for the new approach, the time is almost constant.
These results have verified that the new approach indeed has time complexity of $O(1)$, and the longer the route is, the more time it saves.
On instances in the \textit{wc} set, the  route lengths that occur most frequently are 1{\raise.17ex\hbox{$\scriptstyle\sim$}}10 and 21{\raise.17ex\hbox{$\scriptstyle\sim$}}30, on which the new approach can achieve nearly 1.5$\times$ and 3.5$\times$ speedups over the traditional approach, respectively.
Averaged on all instances in the \textit{wc} set, the new evaluation approach offers nearly 3.3$\times$ speedup over the traditional approach.

\subsection{Sensitivity Analysis of the Population Size \textit{N}}
\label{sec:sensitivity_N}
The population size $N$ is a user-defined parameter that has an important impact on the algorithm's performance.
A larger $N$ directly enables more adequate coverage on the design space of $(\lambda, \gamma)$ in the initialization procedure (see Section~\ref{sec:init}), leading to higher diversity among the initial population and therefore better exploration.
On the other hand, a smaller $N$ facilitates MATE to search more sufficiently in local areas, leading to better exploitation.
Given a limited time budget, the value of $N$ determines the trade-off between exploration and exploitation.
To quantitatively investigate its impact, we tested MATE with different $N$, i.e., 4, 9, 16, 25, 36, 49 and 64, on instances of different scales.
More specifically, we considered all the seven subsets of the \textit{wc} set, and two additional subsets F20{\raise.17ex\hbox{$\scriptstyle\sim$}} (containing four 200-customer instances) and F40{\raise.17ex\hbox{$\scriptstyle\sim$}} (containing four 400-customer instances) of the \textit{jd} set.
For instances in the $\textit{wc}$ set, we set the maximum running time of MATE to 60s; while for instances in F20{\raise.17ex\hbox{$\scriptstyle\sim$}} and F40{\raise.17ex\hbox{$\scriptstyle\sim$}}, we set the maximum running time to 600s.
On each instance, MATE with each considered $N$, denoted as $N_i$, was tested for 30 independent runs, and the average TC, denoted as $TC_i$, was computed.
Then as before, for $N_i$ the PDR against the best-performing $N$ was computed, i.e., $[TC_i-TC_{best}]/TC_{best}$, where $TC_{best}$ is the best performance achieved among all the considered $N$.
Therefore a PDR of 0.00\% means the corresponding $N$ achieves the best performance on the instance.
For each considered $N$, its average PDR on each subset is presented in Table~\ref{tab:analysis_N_results}.
Note the larger the average PDR is, the worse the corresponding $N$ performs on the subset, and an average PDR of 0.00\% means the corresponding $N$ performs the best on every instance in the subset.

One can make two important observations from Table~\ref{tab:analysis_N_results}.
First, on the small-scale instances (the $\textit{wc}_{small}$ subset), the performance of MATE is completely insensitive to the value of $N$.
Second, for instances of medium scales ($M=100$), larger values of $N$ (36, 49 and 64) enabling better exploration perform better.
In particular, each of the three values has achieved the best average PDR on two such subsets, besides Cdp2{\raise.17ex\hbox{$\scriptstyle\sim$}}.
On the other hand, as problem scale increases, the search space grows exponentially.
In this case more sufficient search in local areas is becoming more important under limited time budgets; therefore a smaller value of $N$ (4, 9 and 16) is better.
In summary, the above results show that the parameter $N$ is indeed important for MATE, and setting $N$ to 36 in case of $M \leq 100$, and 16 in case of $M > 100$, is a reasonably good default choice if no other prior knowledge is suggested.

\section{Conclusion}
\label{sec:conclusion}
In this paper, we proposed a memetic algorithm, dubbed MATE, for solving VRPSPDTW.
Compared to existing algorithms, MATE is novel in three aspects: initialization procedure, crossover operator and local search procedure.
Based upon our comprehensive experimental studies, two main conclusions can be drawn.
First of all, MATE is capable of finding better solutions than the state-of-the-art algorithms on a wide range of problem instances.
Notably, MATE finds new best-known solutions on 12 instances in the existing benchmark (65 instances in total).
Second, each novel component integrated into MATE has contributed to its overall performance, and the local search procedure is the one with the biggest contribution.
Moreover, a set of 20 new instances with 200, 400, 600, 800 and 1000 customers were derived from a real-world application of the JD logistics and utilized as a new benchmark set for the large-scale VRPSPDTW.

MATE can be further improved in two aspects.
Currently the maximum length of the subsequences manipulated by the local-search operators is set to 2, which may affect the performance on optimizing lengthy routes.
On the other hand, using a larger maximum length will introduce higher computational costs.
An option is to enhance MATE with self-adaptation that dynamically adjusts the maximum length according to the routes being manipulated.
Another possible improvement to MATE is a more refined scheme of conducting local search.
It has been well recognized in the literature that not all individuals in the population deserve local search.
Hence, mechanisms (e.g., heuristics or learned models) can be integrated into MATE to exclude those 
``unpromising'' individuals from the local search procedure to further reduce the computational costs.

Another future direction is to gradually build MATE into a highly parameterized algorithm framework that supports solving more VRP variants, e.g., multi-depot VRP and VRP with electric fleet.
In this way, one who needs to solve a specific type of VRP can utilize automation tools \cite{LiuTL020,LiuT019,LiuT020,TangLYY21,chen2021new} to search in the configuration space built upon MATE, to obtain an effective algorithm for the specific problem of interest.

\bibliography{mybib}

\end{document}